\pdfoutput=1
\documentclass[11pt]{article}

\usepackage[final]{acl}

\usepackage{float}
\usepackage{stfloats} 

\usepackage{booktabs}
\usepackage{times}
\usepackage{amssymb}
\usepackage{latexsym}
\usepackage[T1]{fontenc}
\usepackage[utf8]{inputenc}
\usepackage{microtype}
\usepackage{inconsolata}
\usepackage{graphicx}
\usepackage{amsmath}
\usepackage{algorithm}
\usepackage{algpseudocode}
\usepackage{lipsum}
\usepackage{multirow}
\usepackage{tabularx}

\title{What to Keep and What to Drop: Adaptive Table Filtering Framework}

\author{
    Wonjune Jang \\
    Department of Mathematics \\
    Myongji University, Korea \\
    \texttt{dnjswnswkd03@mju.ac.kr}
}

\begin{document}
\maketitle

\begin{abstract}
Large language models (LLMs) for table-based reasoning often struggle with large tables due to input length limits. We propose ATF (Adaptive Table Filtering Framework), a modular and question-aware filtering pipeline that prunes uninformative columns and rows using LLM-generated column descriptions, clustering, and sparse-dense alignment scores. ATF integrates seamlessly with existing models (e.g., TAPAS, TAPEX) without retraining. Experiments show that ATF reduces table cells by ~70\%, boosting performance on out-of-domain TableQA tasks while causing slight performance drops on Table Fact Verification, where full-table context is more critical. These results highlight ATF's ability to adaptively balance informativeness and minimalism across tasks. Our code available at: \url{https://github.com/torijune/ATF-Adaptive-Table-Filtering-Framework}
\end{abstract}

\section{Introduction}
Table Question Answering (TableQA) has emerged as a key task in enabling natural language interaction with structured data. Recent advancements increasingly leverage large language models (LLMs ~\cite{openai2024gpt4o}) to interpret semi-structured tables in a flexible, end-to-end fashion~\cite{cheng2023binder, zhang2023reactableenhancingreacttable}. Despite promising progress, current systems continue to face critical limitations, particularly when operating on real-world tables that are lengthy, noisy, or semantically sparse~\cite{zhu2021tatqaquestionansweringbenchmark}.

Most notably, TableQA models often lack mechanisms to selectively compress large or noisy tables, instead encoding the entire table regardless of relevance. This leads to inefficiencies in reasoning and a higher risk of hallucinated or inaccurate answers~\cite{kweon2023openwikitables, wu2024protrix}. Prior attempts to mitigate this include retrieval-based subsetting~\cite{eisenschlos2021mate} and attention-based modeling~\cite{herzig2020tapas}, but these approaches are limited by either weak table-awareness in retrieval or diffuse attention over irrelevant regions. Our proposed framework directly addresses this by incorporating column- and row-level filtering modules that leverage LLMs to assess question-conditioned relevance and explicitly prune irrelevant content prior to answer generation.

Another key challenge is the limited interpretability of filtering decisions. While models like ReActable adopt Chain-of-Thought prompting to express intermediate reasoning ~\cite{zhang2023reactableenhancingreacttable} , they do not offer explicit rationales for the inclusion or exclusion of specific table components. In contrast, our method generates natural language explanations for each column and assigns semantic relevance scores~\cite{glenn2024blendsql}, enabling transparent, user-auditable filtering and enhancing trustworthiness in downstream reasoning.

Furthermore, recent studies have explored the integration of reasoning paths within TableQA pipelines. For example, BINDER~\cite{cheng2023binder} decomposes table-based queries into latent sub-questions to guide multi-hop reasoning, offering structured pathways for inference. Similarly, recent studies have researched utilize question generation and evidence pinpointing to improve faithfulness ~\cite{zhang2023reactableenhancingreacttable}~\cite{seo2024unveilingimplicittableknowledge}. However, even these sophisticated methods rely on fully serialized input tables and do not explicitly address table length constraints or the presence of structural noise. These limitations, as acknowledged by the authors themselves, reduce the robustness and scalability of such systems in complex, real-world environments.

In this paper, we aim to compress the tables using a modular filtering framework that determines \textit{what to keep and what to drop}, retaining only the parts relevant to the question before LLM answers begin. By structurally compressing the table through question-conditioned column and row filters followed by joint filtering we reduce noise, enhance answer accuracy, and support scalable, interpretable TableQA. 

Our column-level filtering leverages LLMs to assign semantic relevance scores to each column. However, a single-pass scoring approach may suffer from instability across multiple runs due to the inherent variability of LLM outputs. For column-level filtering, we propose a robust multipass scoring mechanism that aggregates LLM-generated scores through average- and standard deviation-based LLM scoring and clustering, improving consistency and reliability. For row-level filtering, we adopt a question–cell fusion strategy that combines both sparse retrieval signals (TF-IDF and BM25 ~\cite{jones1972statistical} \cite{robertson1994bm25}) and dense semantic similarities~\cite{reimers2019sbert}, enabling fine-grained and context-aware relevance estimation.

Our system is designed with modular orchestration, making it compatible with easy plug-and-play integration into other frameworks.
Since our filtering module operates entirely prior to model inference, it can be seamlessly integrated with a wide range of downstream TableQA models—including TAPAS, TURL, SQLova, and even SQL-executing prompt-based LLMs—without requiring architectural changes.
This modularity enables flexible deployment and facilitates performance enhancement across different modeling paradigms, regardless of whether they are pre-trained, fine-tuned, or prompt-based.

To evaluate our method in both the in-domain (ID) and out-of-domain (OOD) settings, we conducted experiments using WikiTableQuestions (WTQ) as the ID dataset and Open-WikiTable~\cite{kweon2023openwikitabledatasetopendomain} and AIT-QA~\cite{katsis2021aitqa} as OOD datasets that reflect realistic, noisy, and previously unseen tabular scenarios. The results confirm substantial gains in both performance and robustness, particularly in OOD settings.

The main contributions of this work are as follows:
\begin{itemize}
    \item We propose a novel \textit{pre-processing framework}, ATF (Adaptive Table Filtering), that adaptively prunes table columns and rows based on question relevance, effectively addressing the input token length limitations of Transformer-based TableQA models such as TAPAS and TAPEX.
    
    \item We empirically demonstrate that ATF can reduce table cells by up to 70\% while improving or maintaining performance in out-of-domain TableQA tasks (e.g., Open-WikiTable, AIT-QA), without requiring any additional model fine-tuning or architectural changes.
    
    \item We show that, even with aggressive filtering, ATF preserves sufficient semantics for reasoning, enabling smaller models like TAPAS-base to match or exceed the performance of larger models like TAPAS-large in certain tasks.
    
    \item Our findings also highlight a task-specific trade-off: while ATF improves performance in TableQA—where question-relevant signals are often localized—it may slightly degrade accuracy in Table Fact Verification tasks that require full-table context, revealing the potential for task-aware adaptive filtering strategies.
\end{itemize}

\section{Related Work}
\subsection{Table-Based QA Models}
Recent approaches in TableQA have sought to improve reasoning over structured data by adapting transformer architectures. TAPAS~\cite{herzig2020tapas} introduced a BERT-based model that performs weakly supervised table parsing by selecting relevant cells and learning aggregation operations. While effective, TAPAS encodes the entire table regardless of question relevance, and relies solely on attention mechanisms, which often fail to suppress semantically irrelevant content—particularly in large or noisy tables. MATE~\cite{eisenschlos2021mate} explores a novel multi-view attention mechanism that integrates row-wise, column-wise, and question-based perspectives. This multi-view design helps the model focus more precisely on informative table regions. However, these recent studies do not incorporate explicit table filtering and still process the full table as input. Consequently, these models are prone to content dilution and inefficiencies when applied to real-world tables, which often contain irrelevant information. This not only increases the risk of hallucinations during the QA process but also leads to unnecessarily long input sequences that strain the model’s context capacity.

\subsection{Approaches for Handling Large and Noisy Tables}

To address the challenges posed by lengthy and noisy tables, recent TableQA research has explored various strategies that aim to reduce input complexity prior to LLM-based reasoning. These approaches vary in how they structure the filtering and reasoning pipeline, ranging from symbolic decomposition to dynamic, language-driven table transformation.

One stream of work emphasizes programmatic decomposition, wherein complex questions are broken down into interpretable sub-operations that are mapped to structured queries such as SQL. This strategy allows systems to filter out irrelevant content early in the pipeline, thereby reducing the reasoning burden on LLMs. For instance, frameworks like Plan-of-SQLs (POS)~\citep{nguyen2025interpretablellmbasedtablequestion} achieve strong interpretability and low computational overhead by translating natural language questions into step-wise SQL operations. However, such approaches may struggle with ambiguous inputs or vague semantic signals that are difficult to formalize programmatically.

A complementary direction focuses on dynamic table editing during the reasoning process itself. Rather than pre-filtering the input, models incrementally evolve the table based on intermediate reasoning steps. Techniques such as Chain-of-Table~\citep{wang2024chainoftableevolvingtablesreasoning} refine the table at each stage—by filtering rows, merging cells, or renaming headers—to better align with the evolving semantic context. While this enhances the model’s adaptability, it also introduces potential instability, as each editing operation depends on the accuracy of previous LLM-generated instructions.

Other approaches pursue evidence decomposition alongside question decomposition. For example, DATER~\citep{ye2023largelanguagemodelsversatile} formulates table reasoning as a multi-hop process in which both questions and table evidence are decomposed into smaller units. This allows the model to iteratively gather and validate information in a more interpretable and faithful manner. Nevertheless, such methods typically process the entire table and do not explicitly filter out irrelevant sections, limiting their scalability for large or noisy inputs.

A more recent direction leverages retrieval-augmented generation (RAG) to improve table reasoning scalability. TableRAG~\citep{chen2024tableragmilliontokentableunderstanding} introduces a million-token TableQA framework that retrieves semantically relevant table chunks and fuses them into a generative model using Fusion-in-Decoder. By leveraging a granularity-aware retriever at the row, column, and cell levels, TableRAG supports multi-hop reasoning over extremely large tables. However, its focus is primarily on retrieval coverage and decoding, without fine-grained control over column/row pruning or compatibility with symbolic QA models. Additionally, TableRAG does not offer a standalone filtering module that can be independently adapted or audited in isolation from the LLM backbone.

Another line of research seeks to jointly augment both query and table inputs to facilitate more focused reasoning. ALTER~\citep{zhang-etal-2025-alter} introduces a multi-stage pipeline that applies step-back and sub-query augmentation to the input query, followed by semantic-aware column and row filtering based on LLM-generated auxiliary descriptions. The system then executes SQL queries over the filtered table to complete reasoning tasks. While ALTER effectively improves performance over large tables, its design is tightly coupled with prompt-based SQL execution and zero-shot settings, making it less flexible for integration into fine-tuned transformer-based models.

To overcome these limitations—such as hallucinations and incorrect answers induced by large and noisy tables—our approach introduces a modular filtering framework that operates prior to LLM-based reasoning. By combining LLM-derived semantic relevance scores with sparse and dense retrieval signals, we enable fine-grained, question-conditioned pruning of columns and rows. This design improves robustness, transparency, and computational efficiency in TableQA pipelines. Crucially, our framework is plug-and-play, meaning it can serve as a lightweight front-end to diverse downstream QA models (e.g., TAPAS, TURL, SQLova), offering enhanced interpretability and consistency regardless of backend architecture.

\section{Methodology}
\label{sec:methodology}

We propose \textbf{ATF} (Adaptive Table Filtering), a modular framework designed to address the challenges of large and noisy tables in table-based question answering. Our approach performs interpretable and efficient filtering through a five-stage pipeline that integrates LLM-based semantic reasoning with traditional retrieval-based methods.

\subsection{Motivation}

In Table Question Answering (TableQA), most questions require only a small portion of the table to derive the correct answer. Processing the entire table with large language models (LLMs) is often inefficient and unnecessary, especially when the critical information resides in specific columns and rows that directly align with the question intent. This inefficiency becomes more severe as table size increases, leading to hallucinations or performance drops due to irrelevant noise and limited context windows.

For example, consider the question: “What was Siena like in 2002 when the President was Andrew Jackson?” A reasonable strategy is to first locate the row where the "President" is “Andrew Jackson,” and then extract the corresponding value from the "Siena\_2002" column. In this case, the columns "President" and "Siena\_2002" are essential for answering the question, while others such as "Schl\_1948" or "Aggr\_" are irrelevant and may introduce noise. This highlights the need for a question-aware filtering mechanism that can identify and retain only the relevant columns and discard noisy ones. Furthermore, once the essential columns are identified, the model can focus on extracting the most relevant rows conditioned on the question, thereby reducing the table to a minimal yet sufficient subset of information. 

\begin{figure*}[htbp]
    \centering
    \includegraphics[width=1.0\textwidth]{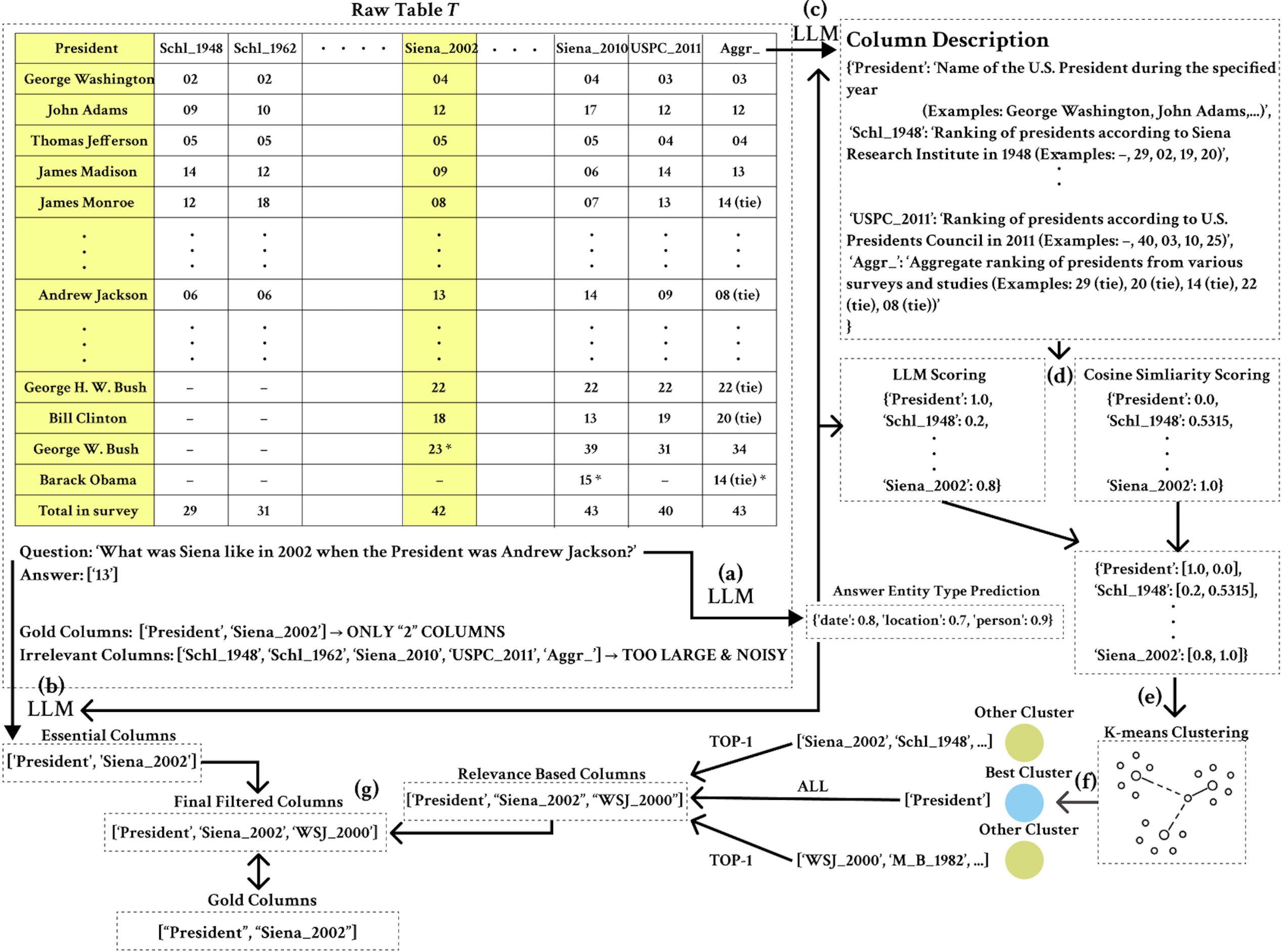}
    \caption{
    Column-level filtering and scoring process for draft data. The figure illustrates the following steps:
    \textbf{(a)} Given the question, LLM predicts the answer entity type as a probability distribution, e.g., \{'date': 0.8, 'location': 0.7, 'person': 0.9\}.
    \textbf{(b)} LLM receives the question and column headers, and predicts which columns are essential for answering the question in a QA context.
    \textbf{(c)} LLM generates natural language descriptions for each column by receiving the column header along with representative cell examples, which are used to assess column relevance.
    \textbf{(d)} Using the generated column descriptions, each column is scored with respect to the question using both LLM-based relevance scoring and embedding-based cosine similarity. The scores are recorded in the form \{'Column\_Name': [LLM Score, Cosine Score]\}.
    \textbf{(e)} K-means clustering is applied to the [LLM Score, Cosine Score] vectors to group columns by importance.
    \textbf{(f)} To select the most QA-relevant cluster, we apply three cluster evaluation metrics, identifying the group that contains the most informative and question-aligned columns. 
    \textbf{(g)} The final set of columns$(C')$ is obtained by merging LLM-predicted essential columns, all columns from the best cluster, and the most relevant column from the other clusters. This reduces 16 columns to 3 while retaining all gold columns.
    }
    \label{fig:column_filtering}
\end{figure*}

To this end, we propose \textbf{ATF} - Adaptive Table Filtering, a modular and interpretable filtering framework that performs column-level and row-level compression based on semantic relevance to the input question, effectively preserving the necessary cell-level content for downstream TableQA reasoning. An overview of the column-level filtering process is presented in Figure~\ref{fig:column_filtering}, while Figure~\ref{fig:row-filtering} illustrates the row-level filtering procedure and the final filtered table.
\subsection{Problem Formulation}

Given a table $T$ with $m$ columns $C = \{c_1, c_2, \ldots, c_m\}$ and $n$ rows $R = \{r_1, r_2, \ldots, r_n\}$, along with a natural language question $q$, our objective is to identify a relevant subset of columns $C' \subseteq C$ and rows $R' \subseteq R$ such that the resulting filtered table $T' = T(C', R')$ maximizes answer accuracy while minimizing processing overhead and hallucination risk. The filtered table $T'$ is then used as input for downstream TableQA models.

\subsection{Answer Entity Type Prediction}

To inform subsequent filtering stages, we first predict the expected \textit{answer entity type} based on the question (Figure~\ref{fig:column_filtering}-(a)). We prompt an LLM with a chain-of-thought reasoning setup to choose from a predefined set of entity types:

\begin{equation}
P(e_i \mid q) = \text{LLM}_{\text{CoT}}(q, \mathcal{E})
\end{equation}

where \\ 
    $\mathcal{E} = \{\text{Person}, \text{Organization}, \ldots, \text{Location}\}$ is the set of possible entity types. The final prediction is:
\[
e^* = \arg\max_{e_i \in \mathcal{E}} P(e_i \mid q)
\]
For example, if the predicted entity type distribution is $P(e_i \mid q)$ = \{'date': 0.8, 'location': 0.7, 'person': 0.9\}, then the most likely entity type $e^*$ is "person". This prediction informs the subsequent column relevance evaluation by enabling entity-aware filtering strategies.

\subsection{Essential Column Extraction}

To ensure semantic completeness and prevent omission of core identifiers (e.g., timestamps, keys), we introduce a dedicated component to extract \textit{essential columns}. Given the question and available table headers, we prompt an LLM to return a conservative subset of columns deemed crucial. The model is instructed to return only a Python list of column names, ensuring compatibility and minimizing hallucinations(Figure~\ref{fig:column_filtering}-(b)). These essential columns are then merged into the final selection set $C'$, regardless of the outcome of the grouping.

We denote the set of essential columns extracted by the LLM as:
\begin{equation}
C_{\text{essential}} = \texttt{LLM}(q, \mathcal{H})
\end{equation}
where $q$ is the question and $\mathcal{H}$ is the set of available column headers.

\subsection{Column-Level Filtering}

We compute column relevance using a hybrid scoring strategy that combines semantic interpretation from LLMs with embedding-based similarity to the question (Figure~\ref{fig:column_filtering}~(c) -- (g)). The detailed algorithm for column-level filtering is described in Appendix~C, Algorithm~\ref{alg:column_filtering}.

\subsubsection{Semantic Description Generation}
Raw column headers and a few sampled cell values often lack sufficient context to support accurate relevance estimation. 
To address this, we prompt the LLM to generate a natural language description $d_i$ that captures the implicit semantics of the column by synthesizing the header $h_i \in \mathcal{H}$ and a representative set of $k$ cell values $V_i = \{v_{i,1}, v_{i,2}, \ldots, v_{i,k}\}$:

\begin{equation}
d_i = \text{LLM}_{\text{desc}}(h_i, V_i)
\end{equation}

This semantic abstraction serves as an intermediate representation that enhances ~\cite{wang2024chainoftableevolvingtablesreasoning} interpretability and provides a more robust foundation for downstream scoring tasks (Figure ~\ref{fig:column_filtering} (c)).

\subsubsection{LLM-based Column Scoring}

Given the inherent stochasticity of large language models, the relevance scores assigned to columns may exhibit variance across multiple executions. To ensure robustness and consistency, we adopt a multi-stage evaluation procedure that aggregates multiple scoring iterations ~\cite{10826109}. This approach enhances both the stability of LLM-based semantic relevance estimation for each column $c_i$ with respect to the input question $q$.

\vspace{0.5em}

\vspace{0.5em}
\noindent\textbf{(1) Iterative Relevance Scoring}
To mitigate noise and non-determinism in LLM predictions, we conduct $N$ scoring iterations per column  ~\cite{10826109}. Each iteration evaluates column relevance by conditioning the LLM on the question $q$, the generated description $d_i$, the predicted answer type $e^*$, and the sampled values $V_i$:

\begin{equation}
s_{i}^{(t)} = \text{LLM}_{\text{score}}^{(t)}(q, d_i, e^*, V_i), \quad t=1,\ldots,N
\end{equation}

Each $s_{i}^{(t)} \in [0, 1]$ reflects the relevance score assigned in iteration $t \leq  N$. Score $1$ means \textit{Essential/Primary key for the answer} and $0$ means \textit{Not relevant at all}

\vspace{0.5em}
\noindent\textbf{(2) Variance-Aware Adjustment}
To penalize inconsistent predictions, we adopt a reliability-aware adjustment that discounts average scores by their associated standard deviation. This ensures that columns which are consistently deemed relevant across multiple LLM calls are preferred over those with high variance, thereby improving the robustness of column selection.

\begin{align}
\mu_i &= \frac{1}{N} \sum_{t=1}^N s_i^{(t)}, \quad
\sigma_i = \sqrt{\frac{1}{N} \sum_{t=1}^N (s_i^{(t)} - \mu_i)^2} \\
\hat{s}_i^{\text{col-final}} &= \mu_i \cdot \left( \frac{1}{1 + \sigma_i} \right)
\end{align}

Here, $\mu_i$ represents the average relevance score for column $c_i$ across $N$ iterations, and $\sigma_i$ captures the standard deviation, indicating the degree of fluctuation in LLM predictions. By scaling the average score $\mu_i$ with the factor $\frac{1}{1 + \sigma_i}$, we effectively downweight columns whose scores vary widely, thereby reducing the influence of unreliable estimates. This formulation can get lower variance and more trustworthy for LLM Scoring \cite{wang2022self}. Consequently, the final score $\hat{s}_i^{\text{col-final}}$ balances both the magnitude and consistency of relevance assessments, favoring columns that are not only relevant but also stably evaluated.

\subsubsection{Embedding-based Similarity}

To complement LLM-based scoring, we compute semantic similarity between the question and each column using sentence-level dense embeddings. Specifically, we concatenate the column header $h_i$ and its generated description $d_i$, and encode both the question $q$ and the column representation using a pretrained sentence embedding model (e.g., \texttt{all-MiniLM-L6-v2}):

\begin{equation}
s_i^{\text{emb}} = \text{cosine\_sim}(\text{embed}(q), \text{embed}(h_i \oplus d_i))
\end{equation}

where $\oplus$ denotes string-level concatenation and \texttt{cosine\_sim} computes the cosine similarity between two embedding vectors. All similarity scores are min-max normalized to $[0,1]$ across all columns to ensure scale invariance.

\vspace{0.5em}
\noindent\textbf{Interpretation}
The final output is a set of relevance scores $\{ \hat{s}_i^{\text{col-final}} \}$ across all columns, combined with the embedding-based scores $\{ s_i^{\text{emb}} \}$ to form a dual-scoring representation. These scores are stored in a dictionary structure of the form \textit{{Column\_Name: [ $\hat{s}_i^{\text{col-final}}, s_i^{\text{emb}}$ ]}}, where each key corresponds to a column and its associated relevance score pair (Figure ~\ref{fig:column_filtering} (d)). The resulting score pairs are subsequently used for column clustering and selection.

\subsection{Column Clustering and Selection}
To capture semantic redundancy and ensure diverse information coverage, we apply clustering-based selection over column representations.

In real-world tables, columns frequently exhibit overlapping semantics or serve analogous roles (e.g., multiple columns related to dates, locations, or categorical identifiers). To address this, we leverage the previously computed dual-scoring dictionary of the form \textit{{Column\_Name: [ $\hat{s}_i^{\text{col-final}}, s_i^{\text{emb}}$ ]}}, which encodes both LLM-derived semantic importance and embedding-based similarity for each column.

Each column is embedded as a 2-dimensional feature vector from this dictionary, and we apply K-means clustering over these vectors to group semantically similar columns.

\subsubsection{Semantic Clustering}

To capture column-level semantic redundancy and structure, we perform \textit{K-means clustering} over the dual-score representations of all columns:

\begin{equation}
\mathcal{C} = \text{K-means}(\{\hat{s}_i^{\text{col-final}}, s_i^{\text{emb}}\}_{i=1}^m)
\end{equation}

The clustering yields a partition $\mathcal{C} = {C_1, C_2, \ldots, C_K}$, where each cluster $C_j$ contains semantically related columns (Figure~\ref{fig:column_filtering} (e)).
The number of clusters $K$ is determined to be 3, based on the Elbow Method and Silhouette Score analysis described in Section~4.8.

\subsection{Cluster Evaluation and Selection Selection}

To determine the optimal set of table columns (Figure ~\ref{fig:column_filtering} (f)), we apply multiple selection strategies that capture different facets of column relevance and diversity. Our approach employs three complementary methods:

\textbf{Semantic Similarity with Cluster Quality:} We compute similarity between the question vector and cluster centroids, weighted by cluster quality metrics that consider both intra-cluster cohesion and inter-cluster separation:
\begin{equation}
\text{Score}(C_j) = \text{sim}(v_q, v_{C_j}) \cdot Q(C_j)
\end{equation}

\textbf{Multi-Criteria Decision Making (MCDM):} This strategy evaluates clusters based on four criteria: relevance score (average column relevance), diversity score (lexical diversity of column names), information density (ratio of high-scoring columns), and size-complexity match (alignment between cluster size and question complexity):
\begin{equation}
\text{Score}(C_j) = w_1 \cdot R_j + w_2 \cdot D_j + w_3 \cdot I_j + w_4 \cdot M_j
\end{equation}

\textbf{Adaptive Confidence Scoring:} We compute confidence scores based on score consistency, strength, and question-type priors, applying dynamic thresholding to filter low-quality clusters:
\begin{equation}
\text{Conf}(C_j) = 0.4 \cdot \text{Cons}_j + 0.4 \cdot \text{Stren}_j + 0.2 \cdot \text{Type}_j
\end{equation}

These methods are ensembled through majority voting: $C^* = \text{mode}(C_1, C_2, C_3)$. In case of disagreement, we select the cluster with the highest individual confidence score. This selection mechanism plays a central role in our framework, as identifying the most relevant cluster is critical for effective filtering. Although the use of multiple selection criteria may appear computationally expensive, each step operates with a time complexity of $\mathcal{O}(m)$, ensuring both efficiency and robustness in the evaluation process. More detail of clustering selection is in Appendix~\ref{appendix:cluster_selection}.

We define the final column set $C'$ as:
\begin{equation}
C' = C_{C^*} \cup \bigcup_{C_j \ne C^*} \text{Top}_1(C_j)
\end{equation}
where $C_{C^*}$ denotes all columns in the selected cluster, and $\text{Top}_1(C_j)$ is the highest-relevance column from non-selected cluster $C_j$.

\subsection{Final Column Selection}
To preserve table semantics, we further include essential columns (e.g., primary keys, timestamps) regardless of cluster assignment:
\begin{equation}
C' \leftarrow C' \cup C_{\text{essential}}
\end{equation}

This ensemble strategy is designed to mitigate the risk of information loss that could arise during the semantic filtering stage. As shown in Figure~\ref{fig:column_filtering}(g), the proposed filtering mechanism effectively retains gold columns required for QA while reducing the total number of columns from 16 to 3 by removing irrelevant and noisy ones.

For example, given the question “What was Siena like in 2002 when the President was Andrew Jackson?”, our method selects only the relevant columns — "President", "Siena\_2002", and "WSJ\_2000" — from the original set of 19 columns, successfully preserving the gold column "President" and "Siena\_2002".

\subsection{Row Ranking and Filtering}

To identify the most relevant rows for the given question, we employ a hybrid retrieval strategy that integrates sparse and dense similarity signals.
Importantly, scoring is performed over the textual content of each row restricted to the previously selected subset of columns, $C'$, rather than using all columns. The detailed algorithm for row-level filtering and final table construction is described in Appendix~C, Algorithm~\ref{alg:row_filtering}.

\begin{figure}[htbp]
    \centering
    \includegraphics[width=\linewidth]{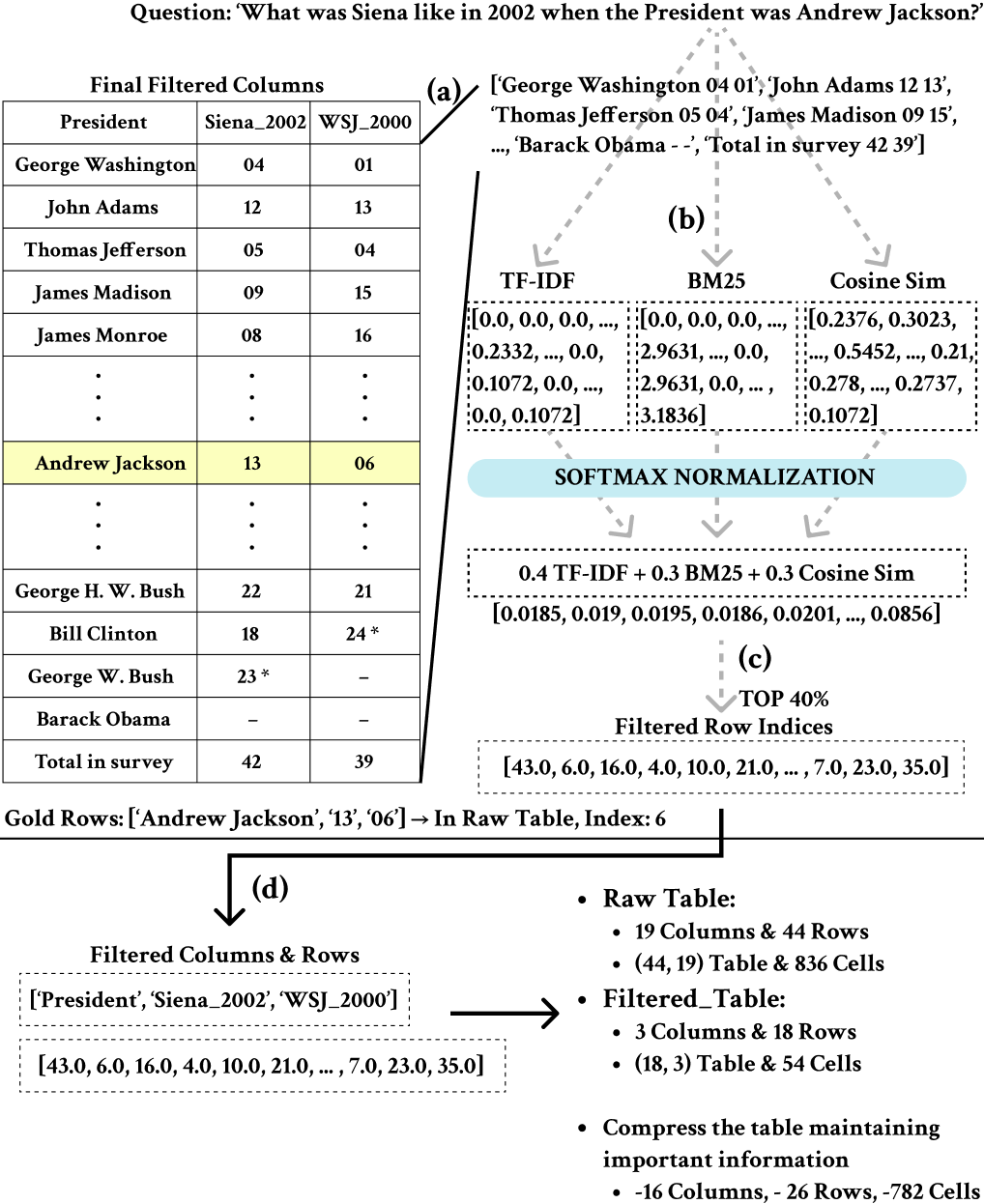}
    \caption{
    Row-wise table compression based on column and row relevance scoring using TF-IDF, BM25, and cosine similarity.  
    \textbf{(a)} The filtered columns from the previous stage (e.g., \textit{President}, \textit{Siena\_2002}, \textit{WSJ\_2000}) are serialized into flattened text strings to represent candidate rows for scoring.  
    \textbf{(b)} Each row is scored with respect to the question using TF-IDF, BM25, and cosine similarity based on textual and embedding-based representations. All scores are normalized using softmax before fusion. The hybrid score is computed as a weighted sum: 0.4 for TF-IDF, 0.3 for BM25, and 0.3 for cosine similarity.  
    \textbf{(c)} The top 40\% most relevant rows are selected based on the hybrid scores.  
    \textbf{(d)} The final compressed table is constructed by intersecting the filtered columns and the top-ranked rows, reducing the full table (44×19) to a smaller table (18×3) while retaining key question-relevant content. This corresponds to a reduction of 16 columns, 26 rows, and 782 cells in total.
    }
    \label{fig:row-filtering}
\end{figure}

\subsubsection{Content Representation}

For each row $r_i$, we generate a flattened text string by concatenating the cell values from the selected columns $C'$ in their original order (Figure~\ref{fig:row-filtering}(a)):

\begin{equation}
\text{text}(r_i) = \bigoplus_{c_j \in C'} T[r_i, c_j]
\end{equation}

For example, if the selected columns are "President", "Siena\_2002", and "WSJ\_2000", the 0-th row \{“President”: “George Washington”, “Siena\_2002”: “04”, “WSJ\_2000”: “01”\} is serialized into the text string: “George Washington 04 01”.

\subsubsection{Hybrid Scoring}

Each row's relevance is scored using three complementary signals in a sparse–dense hybrid manner (Figure~\ref{fig:row-filtering}(b)).  
In TableQA, models often rely on lexical matching against gold columns to retrieve the answer. To better capture both explicit and implicit connections, we adopt a hybrid scoring strategy that combines sparse retrieval methods (TF-IDF and BM25) with dense semantic similarity.  
TF-IDF and BM25 identify direct lexical overlaps between the question and rows, while dense embeddings capture deeper semantic alignment.

\begin{itemize}
  \item \textbf{TF-IDF Similarity} ($s_i^{\text{TF-IDF}}$): Cosine similarity between TF-IDF vectors of the question and row text.
  \item \textbf{BM25 Score} ($s_i^{\text{BM25}}$): Classic token-based ranking using the BM25 retrieval model.
  \item \textbf{Dense Similarity} ($s_i^{\text{dense}}$): Cosine similarity between Sentence-BERT embeddings of the question and row text.
\end{itemize}

\subsubsection{Score Normalization and Fusion}

To address the distributional differences between sparse and dense score vectors, we apply softmax normalization to each scoring signal:

\begin{align}
\tilde{s}_i^{\text{TF-IDF}} &= \text{softmax}(s_i^{\text{TF-IDF}}) \\
\tilde{s}_i^{\text{BM25}} &= \text{softmax}(s_i^{\text{BM25}}) \\
\tilde{s}_i^{\text{dense}} &= \text{softmax}(s_i^{\text{dense}})
\end{align}

We then compute the final fused score:

\begin{equation}
s_i^{\text{row-final}} = 0.4 \cdot \tilde{s}_i^{\text{TF-IDF}} + 0.3 \cdot \tilde{s}_i^{\text{BM25}} + 0.3 \cdot \tilde{s}_i^{\text{dense}}
\end{equation}

These weights were empirically chosen to balance precision and recall across retrieval types.

\subsubsection{Adaptive Row Selection}

To ensure scalability across tables of varying sizes, we adopt an adaptive top-$k$ selection strategy. Given $n$ rows, we select the top-$k$ rows where (Figure ~\ref{fig:row-filtering} (c)):

\begin{equation}
k = \left\lceil \alpha \cdot n \right\rceil
\end{equation}

with $\alpha = 0.4$ by default. This ensures that at least one row is always selected while maintaining efficiency on large tables.

The final selected rows $R'$ are those with the top-$k$ values of $s_i^{\text{row-final}}$. This hybrid, adaptive approach enhances downstream answer generation by reducing noise and emphasizing semantically aligned rows.

\subsection{Final Table Construction}

The filtered table $T'$ is constructed by intersecting selected columns $C'$ and rows $R'$ (Figure ~\ref{fig:row-filtering} (d)):

\begin{equation}
T' = T[R', C']
\end{equation}

This compact representation retains critical information while substantially reducing the input size for downstream QA models.
Given an original table of size $m \times n$ and a compressed table of size $m' \times n'$, the reduction is quantified by $(m \times n) - (m' \times n')$.
For example, a $44 \times 16$ table in Figure~\ref{fig:column_filtering}, ~\ref{fig:row-filtering} can be compressed to $18 \times 3$, reducing the number of processed cells from 836 to 54. This compression improves both input efficiency and quality by focusing the model on the most relevant content.

\section{Experiments}

We conduct extensive experiments to evaluate the effectiveness of ATF on two table-based tasks: Question Answering (QA) and Table Fact Verification (TFV), under two distinct settings: \textit{in-domain} (ID) and \textit{out-of-domain} (OOD).
We define \textbf{in-domain} data as datasets that were used during the fine-tuning of the baseline models, while \textbf{out-of-domain} data refers to datasets that were not observed during this fine-tuning process~\cite{hendrycks2020ood_detection, ben2010domain_adaptation}.

Our central hypothesis is that ATF may degrade performance in ID scenarios, where models are already optimized for the structure and content of raw tables seen during training. In contrast, we expect ATF to show improvements or at least maintain performance in OOD scenarios, where table formats, semantics, and structures deviate significantly from the training distribution.

This hypothesis applies to both QA and TFV tasks. In ID settings, models have likely learned to rely on full table structures as contextual cues, meaning that pruning such tables may result in the loss of relevant information. On the other hand, in OOD settings where tables are often noisy, oversized, or irregular, ATF has the potential to improve generalization by filtering out spurious or redundant content.  

The following experiments are designed to test this hypothesis across both tasks under ID and OOD conditions.

\subsection{Datasets}

We use \textbf{WikiTableQuestions (WTQ)}~\cite{pasupat2015compositionalsemanticparsingsemistructured} as the in-domain dataset for the Table Question Answering (Table QA) task. This dataset was used to fine-tune both of our baseline models and therefore serves as a representative ID setting.

For the Table-based Fact Verification (TFV) task, we use \textbf{TabFact}~\cite{2019TabFactA} as the in-domain dataset. TabFact contains a large number of (table, statement) pairs annotated with binary labels indicating whether the statement can be inferred from the table. As it is also used to fine-tune the baseline TFV models, it represents the ID setting for this task.  
For the out-of-domain evaluation of TFV, we use \textbf{SEM-TAB-FACT}~\cite{wang-etal-2021-semeval}, a challenging benchmark designed to test factual verification over semi-structured tables that differ in style and domain from TabFact. This dataset allows us to evaluate model generalization in TFV beyond the fine-tuning distribution.

For out-of-domain evaluation, we select two datasets for TableQA:  
First, \textbf{Open-WikiTable}~\cite{kweon2023openwikitabledatasetopendomain}, which shares a similar structure with WTQ but requires more complex reasoning, such as multi-row inference and latent column operations.  
Second, we use \textbf{AIT-QA}~\cite{katsis2021aitqa}, a dataset constructed from hierarchical tables in the Airline Industry domain. AIT-QA consists of non-standard, tree-like table structures that challenge traditional TableQA methods.

These datasets allow us to rigorously compare the performance of ATF in both ID and OOD conditions across a range of table complexities and reasoning requirements for both TableQA and TFV tasks.

To investigate the impact of ATF, we apply it to compress input tables prior to inference and analyze its effect on model performance.

\subsection{Baselines}
To evaluate the effectiveness of ATF-based table compression, we select two representative TableQA models as baselines:  
\textbf{TAPAS}~\cite{herzig2020tapas}, which is fintuned by WikiTableQuestion ~\cite{pasupat2015compositionalsemanticparsingsemistructured} and incorporates table position embeddings to filter irrelevant columns implicitly, and  
\textbf{TAPEX}~\cite{liu2022tapextablepretraininglearning}, which is fintuned by WikiTableQuestion and demonstrates strong performance on complex reasoning paths.  
We analyze the performance of both models with and without ATF applied prior to inference.

\subsection{Evaluation Metrics}

To evaluate the performance of table-based Question Answering (QA) models across different datasets—namely \textbf{WikiTableQuestions (WTQ)}, \textbf{Open-WikiTable}, and \textbf{AIT-QA}—we report three metrics: \textbf{Exact Match (EM)}, \textbf{F1 score}, and \textbf{Accuracy (Acc)}. All metrics are computed on normalized textual answers.

\paragraph{Exact Match (EM)}  
EM measures the proportion of predictions that exactly match the ground-truth answers. As a strict metric, EM requires complete agreement between the prediction and the gold answer. To ensure fairness and consistency, we apply a normalization step prior to comparison, which includes lowercasing, punctuation removal, stripping of leading/trailing whitespaces, and normalization of numeric formats (e.g., removing trailing “.0” and numeric commas). Formally, EM is defined as:
\[
\mathrm{EM} = \frac{1}{N} \sum_{i=1}^{N} \mathbb{1}(\hat{y}_i = y_i)
\]
where \( \hat{y}_i \) and \( y_i \) denote the normalized predicted and gold answers for instance \( i \), and \( \mathbb{1}(\cdot) \) is the indicator function.

\paragraph{F1 Score}  
The F1 score captures partial correctness by computing the harmonic mean of precision and recall based on token overlap between the predicted and gold answers. This metric is particularly effective when answers consist of multiple tokens, as it rewards partial matches. Let \( \hat{y} \) and \( y \) represent the sets of tokens in the normalized predicted and gold answers respectively. Then:
\[
\mathrm{Precision} = \frac{|\hat{y} \cap y|}{|\hat{y}|}, \quad
\mathrm{Recall} = \frac{|\hat{y} \cap y|}{|y|}
\]
\[
\mathrm{F1} = \frac{2 \cdot \mathrm{Precision} \cdot \mathrm{Recall}}{\mathrm{Precision} + \mathrm{Recall}}
\]
If there is no token overlap (i.e., \( |\hat{y} \cap y| = 0 \)), the F1 score is set to zero.

\paragraph{Accuracy (Acc)}  
Accuracy is used primarily for classification-based tasks, such as Table Fact Verification (TFV). It measures the proportion of examples for which the model predicts the correct class label. Let \( \hat{y}_i \in \{0, 1\} \) be the predicted label and \( y_i \) be the gold label:
\[
\mathrm{Accuracy} = \frac{1}{N} \sum_{i=1}^{N} \mathbb{1}(\hat{y}_i = y_i)
\]
For 2-class settings (e.g., Entailed vs. Refuted), this metric aligns directly with standard classification accuracy. For multi-class tasks (e.g., Entailed, Refuted, Not Enough Info), macro-averaged accuracy may also be reported where appropriate.

\paragraph{Normalization Procedure}  
To standardize evaluations across datasets, we apply the following normalization steps prior to computing all metrics:
\begin{itemize}
    \item Convert text to lowercase and remove leading/trailing whitespace.
    \item Remove commas in numerical values (e.g., \texttt{"11,327"} → \texttt{"11327"}).
    \item Convert floating integers to integer form (e.g., \texttt{"11327.0"} → \texttt{"11327"}).
    \item Strip punctuation where applicable.
\end{itemize}

\subsection{Results}

\begin{table}[t]
\centering
\caption{TableQA In-domain (ID) summarization results on WikiTableQuestions of TAPAS, TAPEX.}
\label{tab:in_domain_tableQA}
\begin{tabularx}{\linewidth}{lccc}
\toprule
\multicolumn{4}{c}{\textbf{In-domain: WikiTableQuestions (TableQA)}} \\
\midrule
\textbf{Model} & \textbf{EM} & \textbf{F1} & \textbf{Cell↓} \\
\midrule
\textbf{TAPAS}{\small\textbf{-base}} & 0.296 & 0.313 & - \\
\textbf{TAPAS}{\small\textbf{-base + ATF}} & \textbf{0.189} & \textbf{0.203} & \textbf{↓68.2\%} \\
\midrule
\textbf{TAPEX} & 0.414 & 0.416 & - \\
\textbf{TAPEX} {\small + \textbf{ATF}} & \textbf{0.267} & \textbf{0.268} & \textbf{↓68.2\%} \\
\bottomrule
\end{tabularx}
\end{table}

\begin{table*}[!t]
\centering
\caption{TableQA Out-of-domain(OOD) summarization results on Open-WikiTableQuestion and AIT-QA of TAPAS, TAPEX.}
\label{tab:ood_results}
\resizebox{\textwidth}{!}{%
\begin{tabular}{lcccccc}
\toprule
& \multicolumn{6}{c}{\textbf{Out-of-domain}} \\
\cmidrule(lr){2-7}
\textbf{Model} 
& \multicolumn{3}{c}{\textbf{Open-WikiTable}} 
& \multicolumn{3}{c}{\textbf{AIT-QA}} \\
\cmidrule(lr){2-4} \cmidrule(lr){5-7}
& \textbf{EM} & \textbf{F1} & \textbf{Cell↓} 
& \textbf{EM} & \textbf{F1} & \textbf{Cell↓} \\
\midrule
\textbf{TAPAS} & 0.599 & 0.619 & - & 0.376 &  0.409 & - \\
\textbf{TAPAS}{\small + \textbf{ATF}} & \textbf{0.610 (↑1.8\%)} & \textbf{0.627 (↑1.3\%)} & \textbf{↓69.9\%} & \textbf{0.508(↑35.4\%)} & \textbf{0.535 (↑30.8\%)} & \textbf{↓67.0\%} \\
\midrule
\textbf{TAPEX} & 0.446 & 0.450 & - & 0.369 & 0.369 & - \\
\textbf{TAPEX}{\small + \textbf{ATF}} & \textbf{0.489 (↑9.6\%)} & \textbf{0.494 (↑9.8\%)} & \textbf{↓69.9\%} & \textbf{0.375 (↑1.4\%)} & \textbf{0.375 (↑1.4\%)} & \textbf{↓67.0\%} \\
\bottomrule
\end{tabular}%
}
\end{table*}

\begin{table*}[t]
\centering
\caption{Table Fact Verification (TFV) Out-of-domain (OOD) summarization results comparing TAPAS-base and TAPAS-large on SEM-TAB-FACT.}
\label{tab:TFV-ID_OOD setting results}
\resizebox{\textwidth}{!}{%
\begin{tabular}{lcccccc}
\toprule
\multirow{3}{*}{\textbf{Model}} & \multicolumn{2}{c}{\textbf{In-domain: TabFact}} & \multicolumn{4}{c}{\textbf{Out-of-domain: SEM-TAB-FACT}} \\
\cmidrule(lr){2-3} \cmidrule(lr){4-7}
& \multicolumn{2}{c}{\textbf{Test}} & \multicolumn{2}{c}{\textbf{Dev}} & \multicolumn{2}{c}{\textbf{Test\_a}} \\
& \textbf{Acc} & \textbf{Cell↓} & \textbf{Acc} & \textbf{Cell↓} & \textbf{Acc} & \textbf{Cell↓} \\
\midrule
\textbf{TAPAS{\small - base}} & 0.771 & - & 0.706 & - & 0.730 & - \\
\textbf{TAPAS{\small -base + ATF}} & 0.760 (↓1.4\%) & ↓71.5\% & 0.683 (↓3.3\%) & ↓69.9\% & 0.703 (↓3.7\%) & ↓67.0\% \\
\midrule
\textbf{TAPAS{\small - large}} & 0.797 & - & 0.734 & - & 0.751 & - \\
\textbf{TAPAS{\small -large + ATF}} & 0.787 (↓1.3\%) & ↓71.5\% & 0.721 (↓1.8\%) & ↓69.9\% & 0.695 (↓7.5\%) & ↓67.0\% \\
\bottomrule
\end{tabular}%
}
\end{table*}

To investigate the generalizability of ATF, we conduct experiments under two separate settings: \textbf{in-domain (ID)} and \textbf{out-of-domain (OOD)}. We define the in-domain setting as evaluation on datasets used for fine-tuning the base TableQA models (e.g., WikiTableQuestions), and the out-of-domain setting as evaluation on unseen datasets that were not exposed during training (e.g., Open-WikiTable, AIT-QA).

Interestingly, contrary to our initial assumption that ATF would consistently improve performance by removing irrelevant table content, our results reveal a more nuanced pattern. As shown in Table~\ref{tab:in_domain_tableQA}, ATF significantly degraded performance on the in-domain dataset. We attribute this to the fact that base models were fine-tuned using \textit{Raw tables, unfiltered tables}, meaning they have adapted to the full structure and context. When ATF reduces cells from already familiar and clean tables, it may remove relevant context, leading to performance drops in EM and F1.

In contrast, as shown in Table~\ref{tab:ood_results}, ATF significantly boosts performance in the out-of-domain setting. On datasets like Open-WikiTable and AIT-QA, both TAPAS and TAPEX benefit from ATF with substantial gains in both EM and F1. This is particularly noteworthy given the characteristics of these datasets: Open-WikiTable requires more complex reasoning compared to WikiTableQuestions, and AIT-QA contains hierarchical and structurally complex tables, which differ greatly from standard flat tables. Despite these challenges, ATF enhances performance by removing noisy or redundant content, helping the model focus on core information.

These findings suggest that while ATF may be suboptimal for in-domain datasets that share distributional similarity with the fine-tuning data, it plays a critical role in improving robustness and generalization under distribution shift. Therefore, ATF is especially promising for real-world applications where tables are noisy, large, or structurally unfamiliar.

\begin{table}[t]
\centering
\caption{Comparison of TAPAS-base and TAPAS-large (fine-tuned on WTQ) on AIT-QA (EM / F1)}
\label{tab:tapas_ATF_vs_large}
\begin{tabular}{lcc}
\toprule
\multicolumn{3}{c}{Dataset: \textbf{AIT-QA}} \\
\midrule
\textbf{Model} & \textbf{EM} & \textbf{F1} \\
\midrule
\textbf{TAPAS}{\small \textbf{-base}} (110M) + ATF  & \textbf{0.508} & 0.535 \\
\textbf{TAPAS}{\small \textbf{-large}} (343M)        & 0.505 & \textbf{0.561} \\
\bottomrule
\end{tabular}
\end{table}

We compare TAPAS-base with ATF against a raw TAPAS-large model on the out-of-distribution (OOD) benchmark AIT-QA. As shown in Table~\ref{tab:tapas_ATF_vs_large}, TAPAS-base with ATF slightly outperforms TAPAS-large in terms of exact match (EM), while TAPAS-large achieves higher F1 scores.
These results suggest that ATF is effective in improving answer accuracy (EM) even when applied to smaller models with fewer parameters.
This highlights the potential of ATF as a lightweight preprocessing technique for enhancing domain generalization performance without relying on large-scale models.

We further investigate the impact of ATF on the Table Fact Verification (TFV) task using the TabFact (ID) and SEM-TAB-FACT (OOD) datasets. As summarized in Table~\ref{tab:TFV-ID_OOD setting results}, we observe a consistent degradation in accuracy when applying ATF across both TAPAS-base and TAPAS-large models.

On the in-domain TabFact test set, ATF slightly reduces accuracy by 1.4\% for TAPAS-base and 1.3\% for TAPAS-large, despite a substantial 71.5\% reduction in table cells. This suggests that in TFV tasks, which often require fine-grained, holistic reasoning over the full table context, aggressive cell pruning may remove critical evidence required for correct entailment prediction.

The trend becomes more pronounced in the out-of-domain SEM-TAB-FACT benchmark. On the dev and test\_a splits, TAPAS-base + ATF shows drops of 3.3\% and 3.7\% in accuracy respectively, while TAPAS-large + ATF suffers 1.8\% and 7.5\% accuracy drops. Notably, these performance declines occur despite consistent reductions in cell count (↓69.9\% and ↓67.0\%), indicating that ATF's compression comes at the cost of essential contextual information in TFV settings.

These findings underscore a critical distinction between TableQA and TFV: while the former benefits from targeted cell filtering to reduce noise, the latter relies on comprehensive table semantics. As such, static compression techniques like ATF, though effective for QA, may be suboptimal for verification tasks requiring global reasoning. This highlights the need for \textbf{task-adaptive filtering mechanisms} that account for the nature of the downstream task.

\subsection{Ablation Study}

ATF consists of two key symbolic filtering stages: column-level and row-level filtering.  
Column-level filtering aims to remove attributes irrelevant to the question (e.g., unrelated metadata), while row-level filtering focuses on eliminating entries that, despite containing relevant columns, do not provide the answer to the question (e.g., distractor rows).

\begin{table*}[!t]
\centering
\caption{Ablation results of ATF on Open-WikiTableQuestion.}
\label{tab:row_column_ablation}
\begin{tabular}{lccc}
\toprule
\textbf{Model} & \textbf{EM} & \textbf{F1} & \textbf{Cell Reduction} \\
\midrule
\textbf{TAPAS}{\small -base-wtq} & 0.599 & 0.619 & - \\
\quad \textbf{(i)} w/o column filtering & 0.606 (↓ 0.7\%) & 0.621 (↓ 1.0\%) & ↓ 56.5\% \\
\quad \textbf{(ii)} w/o row filtering & 0.590 \textbf{(↓ 3.3\%)} & 0.612 (↓ 2.4\%) & ↓ 30.7\% \\
\quad \textbf{(iii)} \textbf{ATF (Ours)} & \textbf{0.610} & \textbf{0.627} & \textbf{↓ 69.9\%} \\
\addlinespace
\textbf{TAPEX}{\small -base-wtq} & 0.446 & 0.450 & - \\
\quad \textbf{(i)} w/o column filtering & 0.470 \textbf{(↓ 3.9\%)} & 0.475 (↓ 3.8\%) & ↓ 56.5\% \\
\quad \textbf{(ii)} w/o row filtering & 0.473 (↓ 3.3\%) & 0.478 (↓ 3.2\%) & ↓ 30.7\% \\
\quad \textbf{(iii)} \textbf{ATF (Ours)} & \textbf{0.489} & \textbf{0.494} & \textbf{↓ 69.9\%} \\
\bottomrule
\end{tabular}
\end{table*}

To analyze the contribution of each filtering component, we conduct an ablation study comparing three variants:  
\textbf{(i)} without column filtering,  
\textbf{(ii)} without row filtering,  
and \textbf{(iii)} the full ATF pipeline.  

As shown in Table~\ref{tab:row_column_ablation}, both TAPAS and TAPEX experience performance degradation when either the column or row filtering component is removed.  
Interestingly, in the case of TAPAS, disabling row-level filtering even led to lower accuracy than the raw model, highlighting the importance of row pruning in reducing distractive context.

These findings confirm that both column- and row-level filtering stages are complementary and jointly contribute to performance gains.  
Moreover, ATF consistently achieves the highest cell reduction rate (↑ 69.9\%), demonstrating its effectiveness in compressing input size while improving or maintaining accuracy.

A more detailed analysis of the reduction at column, row, cell, and token level of ATF will be presented in Section 4.7.

\subsubsection{Clustering Selection vs Top-K Selection}

ATF adopts a clustering-based approach for column selection, where each column is represented as a 2-dimensional vector $\texttt{[}\hat{s}_i^{\text{col-final}},\ s_i^{\text{emb}}\texttt{]}$ and grouped using K-means clustering. A three-stage clustering evaluation process is then applied to choose the final relevant columns. While this method offers better semantic grouping, it incurs additional computational cost compared to simpler strategies.

To evaluate the effectiveness of clustering-based selection, we compare ATF’s K-means-based approach against a simpler L2-norm-based Top-K selection method. In the Top-K approach, columns are ranked by their L2 norm scores $\|\mathbf{v}_i\|_2 = \sqrt{(\hat{s}_i^{\text{col-final}})^2 + (s_i^{\text{emb}})^2}$, and the top-ranked ones are selected. To ensure fairness, we align the selection ratio: when the number of raw columns is less than 10, we select the top 3; otherwise, we select the top 40\%.

\begin{table}[htbp]
\centering
\caption{Comparison of K-means Clustering vs. Top-K Selection (TAPAS/TAPEX on Open-WikiTable)}
\label{tab:KMenas_TopK_Ablation}
\begin{tabular}{lcc}
\toprule
\textbf{Model} & \textbf{EM} & \textbf{F1} \\
\midrule
\textbf{TAPAS}{\small -base-wtq} & & \\
\quad K-means (Ours) & \textbf{0.610} & \textbf{0.627} \\
\quad Top-K & 0.561 \textbf{(↓ 8.0\%)} & 0.582 \\
\addlinespace
\textbf{TAPEX}{\small -base-wtq} & & \\
\quad K-means (Ours) & \textbf{0.489} & \textbf{0.494} \\
\quad Top-K & 0.461 (↓ 5.7\%) & 0.466 \\
\bottomrule
\end{tabular}
\end{table}

As shown in Table~\ref{tab:KMenas_TopK_Ablation}, both TAPAS and TAPEX models show a significant drop in EM and F1 scores—over 5\%—when replacing the clustering-based selection with simple Top-K filtering. This suggests that ATF’s clustering mechanism, despite its higher complexity, is more effective at identifying question-relevant columns than naïve score-based selection. The ability to model fine-grained semantic groupings contributes to improved question-answering performance.

\subsection{Filtering Efficiency Analysis}
We evaluate the efficiency of ATF's column filtering by examining how many columns are removed for tables with varying raw column counts, as well as how the reduction ratio influences the distribution of those tables.

\subsubsection{Column Reduction Analysis}

\begin{figure}[htbp]
    \centering
    \includegraphics[width=\linewidth]{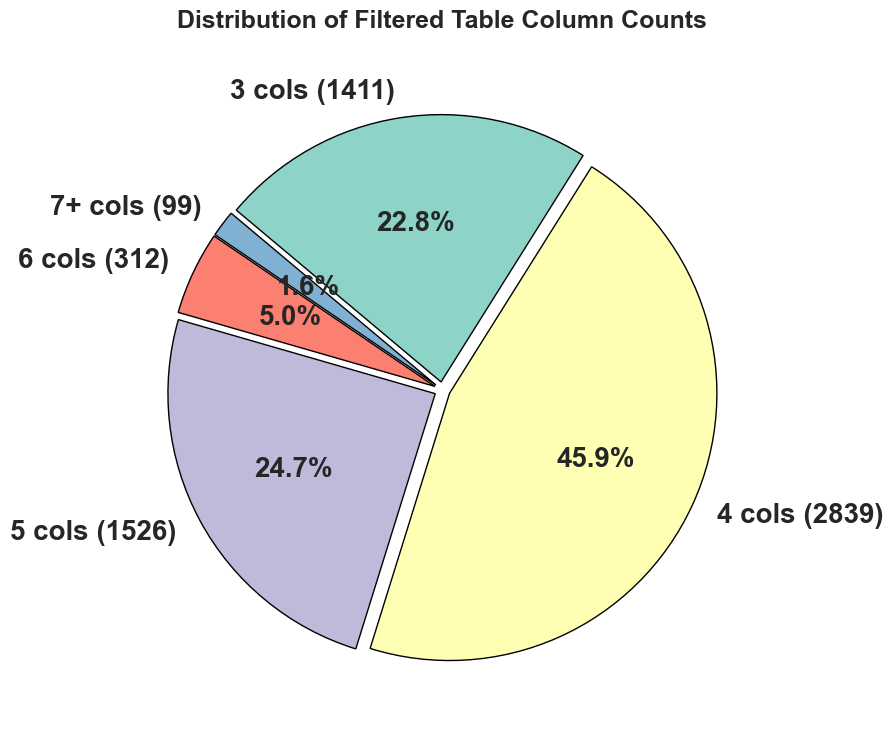}
    \caption{Column Reduction Ratio}
    \label{fig:column_reduction_count_pie}
\end{figure}

We analyze the reduction ratio of columns. Figure~\ref{fig:column_reduction_count_pie} illustrates the distribution of column reduction counts across all raw tables.

We observe that a significant portion (45.9\%) of raw tables are filtered down to 4 columns. The next most common outcomes are 5 columns (24.7\%) and 3 columns (22.8\%). This suggests that the number of columns necessary for TableQA tasks is typically around 3 to 5.
These results suggest that column filtering has limited impact on raw tables with 3 to 5 columns, as most of them are either essential or the table is too small to be further reduced.

\begin{figure*}[t]
    \centering
    \includegraphics[width=1.0\textwidth]{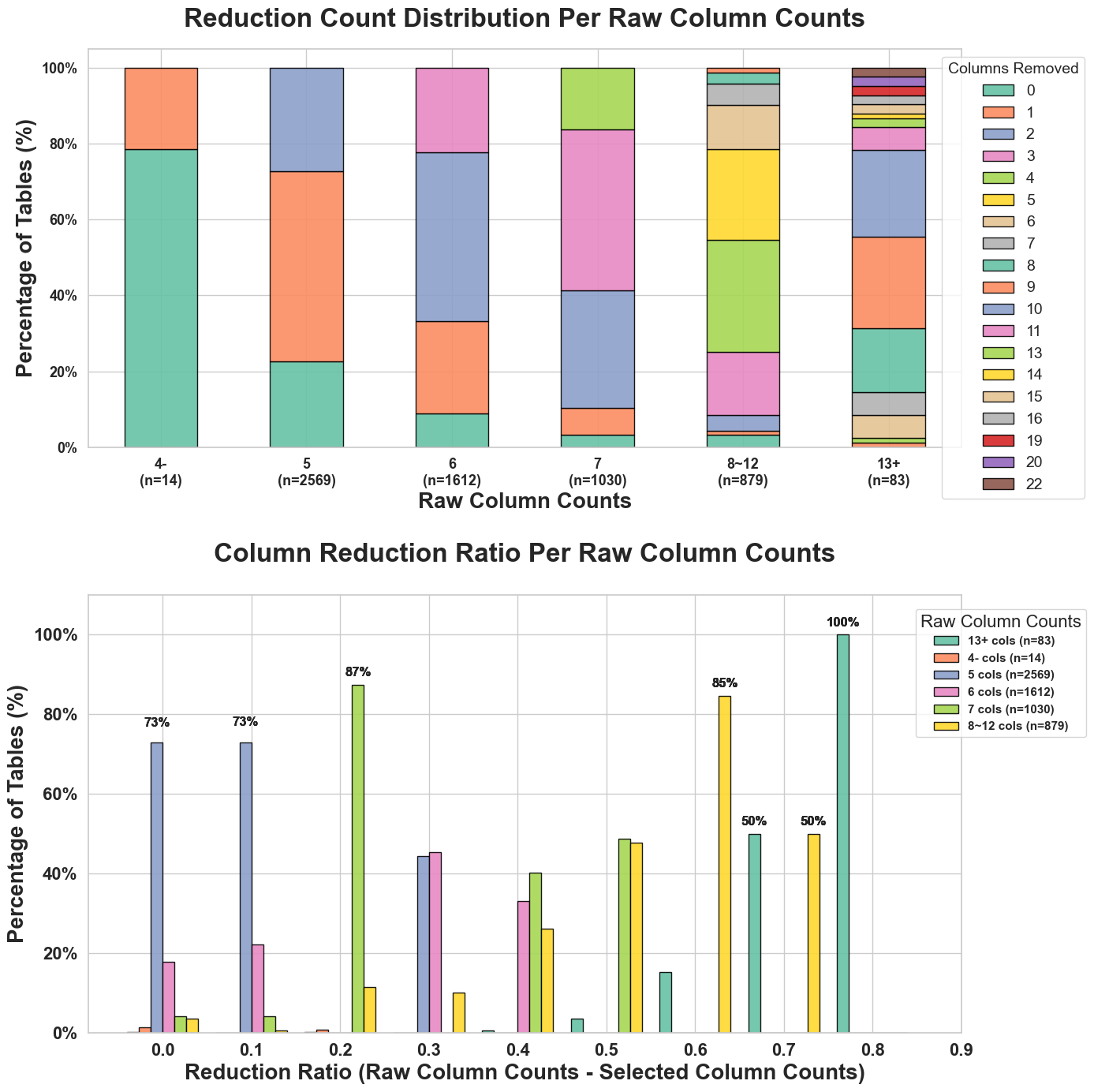}
    \caption{Visualization of how ATF reduces columns: (Top) Distribution of columns removed across raw column count groups, (Bottom) Reduction ratio distribution by group.}
    \label{fig:ColumnReduction_Subplot}
\end{figure*}

To further investigate the relationship between the number of raw columns and the extent of filtering, we analyze the distribution of column reductions based on the raw column counts of the tables Figure~\ref{fig:ColumnReduction_Subplot}.

Figure~\ref{fig:ColumnReduction_Subplot} illustrates the effectiveness of the proposed column filtering method in terms of both absolute reduction count per raw column counts (top) and normalized reduction ratios per raw column count (bottom).

As shown in Figure~\ref{fig:ColumnReduction_Subplot} (Top), when the number of raw columns is four or fewer("4-"), most columns are likely essential for QA, leading to minimal filtering (0–1 columns removed). In contrast, for tables with 6 to 7 columns—the most frequent cases—ATF typically filters out 2 to 3 columns, indicating a meaningful reduction. For wider tables with 8–12 or more than 13 columns, ATF frequently removes between 7 and 22 columns, demonstrating its effectiveness in reducing noise in complex table structures.

As illustrated in Figure~\ref{fig:ColumnReduction_Subplot} (Bottom), Consistent with previous observations(Figure~\ref{fig:ColumnReduction_Subplot} (Top)), tables with fewer columns (e.g., $\leq$ 5) tend to have lower reduction ratios (0.0–0.3), as most columns are essential for QA. For tables with 5–7 columns, the most common cases, the reduction ratio is typically distributed around 0.2–0.5, indicating moderate filtering. In contrast, tables with 8–12 or 13+ columns show higher reduction ratios (0.6–0.8), suggesting that ATF effectively filters out non-essential columns in wider and noisier tables.

\subsubsection{Row Reduction Analysis}

To evaluate the effectiveness of row-level filtering, we analyze the number and proportion of rows removed during the table pruning process.

\begin{figure}[htbp]
    \centering
    \includegraphics[width=\linewidth]{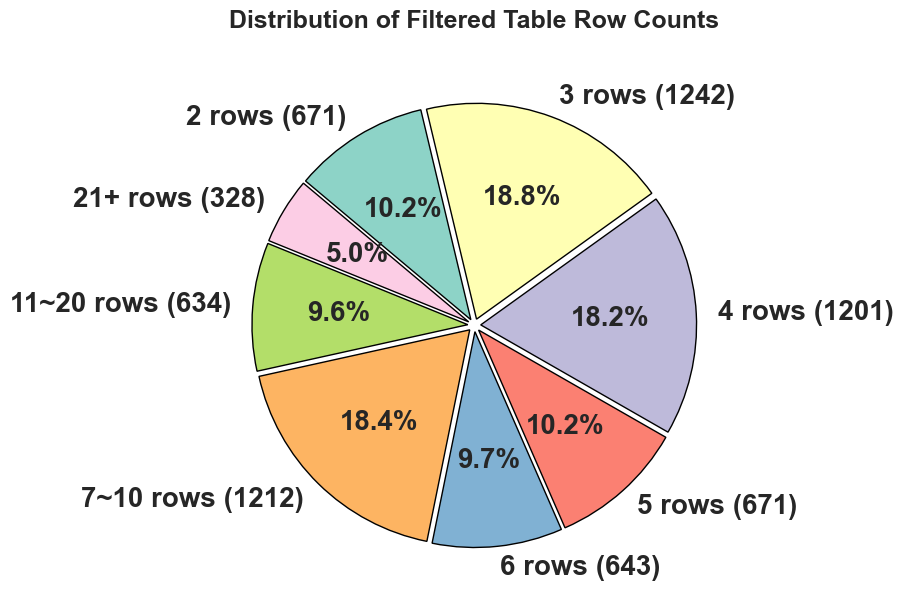}
    \caption{Distribution of filtered row counts across tables.}
    \label{fig:row_reduction_count_pie}
\end{figure}

As shown in Figure~\ref{fig:row_reduction_count_pie}, the majority of filtered tables retain only 2 to 5 rows. For tables with larger raw row counts, the number of retained rows tends to increase accordingly, as ATF performs reduction in a ratio-based manner.

\begin{figure}[htbp]
    \centering
    \includegraphics[width=0.9\linewidth]{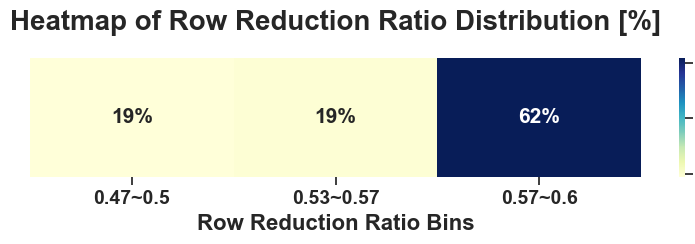}
    \caption{Heatmap of row reduction ratio distribution across all tables.}
    \label{fig:row_reduction_ratio_heatmap}
\end{figure}

Figure~\ref{fig:row_reduction_ratio_heatmap} shows the distribution of row reduction ratios. Notably, 62\% of tables fall within the 0.57--0.6 range, indicating that ATF significantly reduces table size while preserving essential content.

\subsubsection{Cell \& Token Reduction Analysis}

To evaluate the overall impact of ATF on table size, we analyze the reduction ratio of table content from two perspectives: (1) the cell-level in DataFrame format, and (2) the token-level in linearized text format.

\begin{figure}[htbp]
    \centering
    \includegraphics[width=1.0\linewidth]{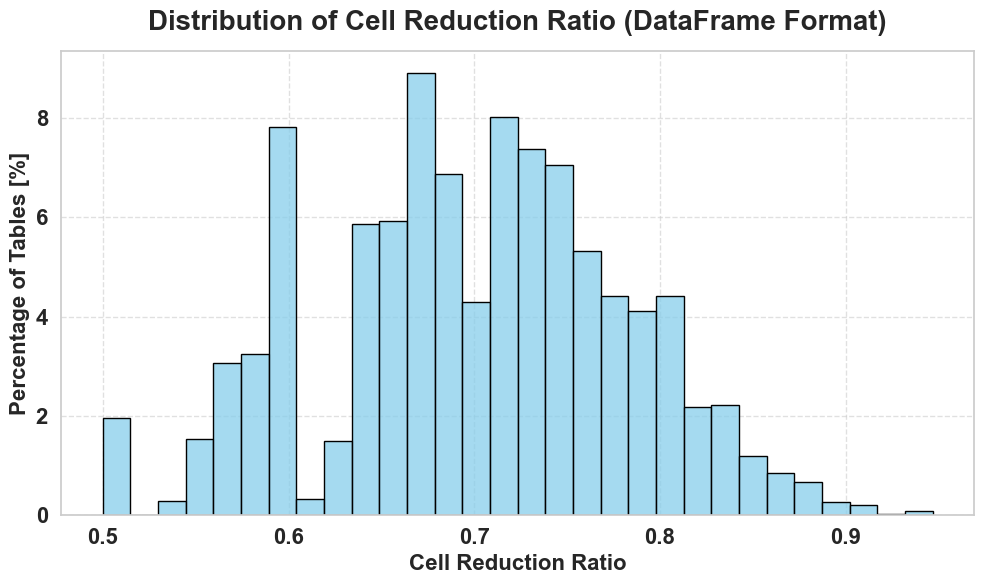}
    \caption{Distribution of cell-level reduction ratios across filtered tables.}
    \label{fig:cell_reduction}
\end{figure}

As shown in Figure~\ref{fig:cell_reduction}, a significant number of filtered tables exhibit cell reduction ratios between 0.5 and 0.9. This trend is consistent with our earlier findings that tables with more rows and columns tend to contain more noisy cells, which are effectively removed by ATF.

\begin{figure}[htbp]
    \centering
    \includegraphics[width=1.0\linewidth]{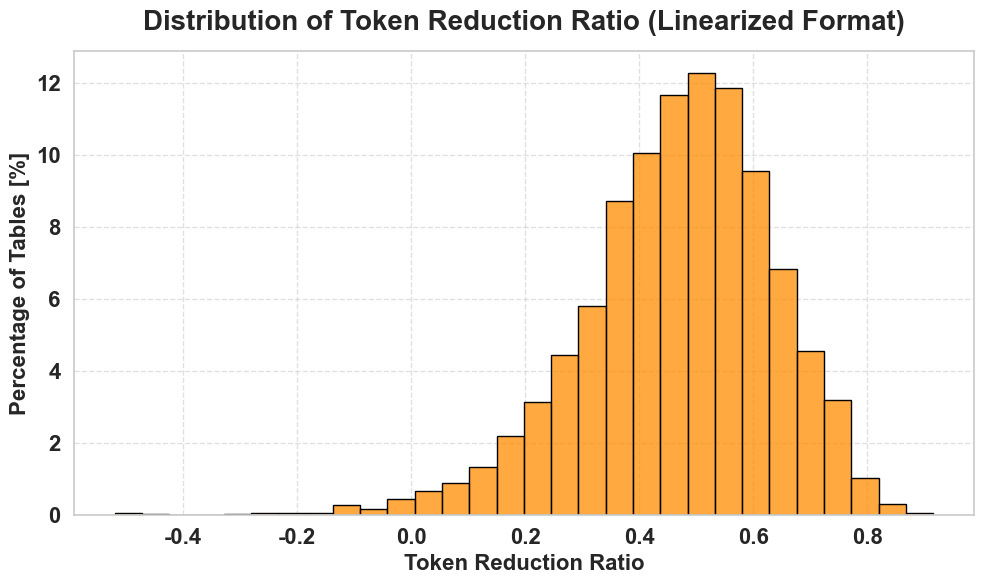}
    \caption{Distribution of token-level reduction ratios after linearization.}
    \label{fig:token_reduction}
\end{figure}

Since input tables must be linearized into plain text for LLMs, we also analyze token-level reduction ratios. As shown in Figure~\ref{fig:token_reduction}, most filtered tables achieve token reductions between 30\% and 70\%. This demonstrates that ATF effectively compresses verbose table structures into a more compact format, enabling compatibility with the input token limits of LLMs ~\cite{vaswani2017attention}.

\subsubsection{Effect of Input Length Constraint}

Transformer-based TableQA models like TAPAS and TAPEX inherit token length limits from their base architectures—512 for TAPAS (BERT~\cite{devlin2019bertpretrainingdeepbidirectional}) and 1024 for TAPEX (BART~\cite{lewis-etal-2020-bart}). Large or noisy tables often exceed these limits, leading to truncation and potential information loss.

To evaluate this, we measured how often inputs exceeded length constraints before and after ATF filtering. As shown in Table~\ref{tab:combined_length_overflow_ratio}, ATF dramatically reduces overflow cases across all datasets. For example, in Open-WikiTable, the overflow ratio for TAPAS dropped from 22.2\% to 2.4\%, and for TAPEX from 7.8\% to 0.5\%.

This suggests that ATF improves not only generalization but also input compatibility, particularly for long or unstructured tables.

\begin{table*}[t]
\centering
\caption{Token length overflow ratio before and after applying ATF across all TableQA and Table Fact Verification (TFV) datasets.}
\label{tab:combined_length_overflow_ratio}
\begin{tabular}{llccc}
\toprule
\textbf{Task} & \textbf{Model} & \textbf{Total} & \textbf{Raw \%} & \textbf{ATF \%} \\
\midrule
TableQA (WTQ - Test) & TAPAS & 4258 & 35.9\% & 5.4\% \\
                     & TAPEX & 4258 & 12.4\% & 2.2\% \\
\midrule
TableQA (Open-WikiTable - Test) & TAPAS & 6602 & 22.2\% & 2.4\% \\
                                & TAPEX & 6602 & 7.8\% & 0.5\% \\
\midrule
TableQA (AIT-QA - Test) & TAPAS & 459 & 21.8\% & 2.6\% \\
                        & TAPEX & 459 & 7.6\% & 0.0\% \\
\midrule
TFV (TabFact - Test) & TAPAS & 12739 & 16.7\% & 0.01\% \\
\midrule
TFV (SEM-TAB-FACT - Dev) & TAPAS & 463 & 25.5\% & 1.1\% \\
\midrule
TFV (SEM-TAB-FACT - Test\_a) & TAPAS & 522 & 13.4\% & 0.0\% \\
\bottomrule
\end{tabular}
\end{table*}

\subsection{K-Means Cluster Size Selection}
In ATF, K-means clustering is applied to partition columns based on their question relevance scores. The number of clusters $K$ is a critical hyperparameter that directly influences the effectiveness of column filtering and, consequently, the overall QA performance.

\begin{figure}[htbp]
    \centering
    \includegraphics[width=1.0\linewidth]{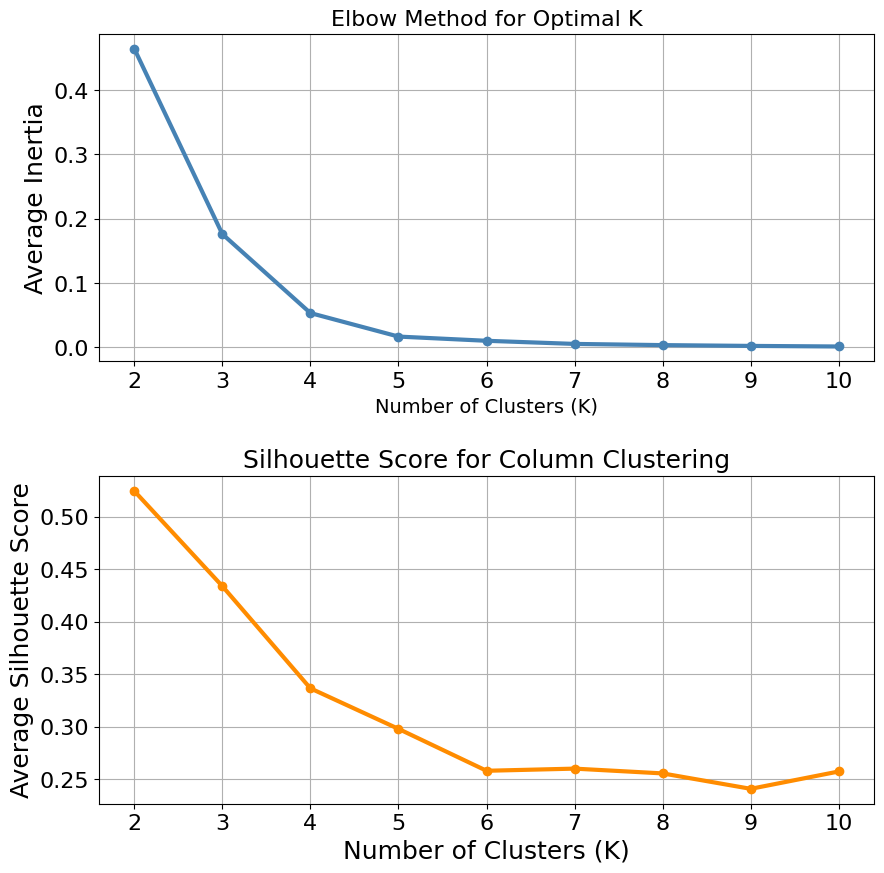}
    \caption{Evaluation of cluster size using Elbow Method (Top) and Silhouette Score (Bottom).}
    \label{fig:kmeans_analysis}
\end{figure}

To determine the optimal value of $K$, we conducted an empirical analysis using the Elbow Method ~\cite{james2013introduction} and Silhouette Score ~\cite{rousseeuw1987silhouettes} (Figure~\ref{fig:kmeans_analysis}). The Elbow plot (Top) shows that the average inertia decreases sharply up to $K=3$ and flattens thereafter, suggesting $K=3$ or $K=4$ as potential candidates.

To further refine the choice, we examined the Silhouette Score Figure~\ref{fig:ColumnReduction_Subplot} (Bottom). While $K=2$ yields the highest score, $K=3$ maintains a notably higher score than $K=4$, suggesting better cluster cohesion and separation. In addition, as shown in Figure~\ref{fig:ColumnReduction_Subplot} (Top), most raw tables contain only 5–7 columns, implying that setting $K \geq 4$ may result in overly granular clustering where some clusters contain only a single column. Based on this combined evidence, we set the number of clusters in ATF to $K=3$.

\section{Conclusion}
We introduce a modular filtering framework for TableQA that improves both precision and efficiency by explicitly identifying what to keep and what to drop. Our method, ATF, is lightweight and model-agnostic, making it broadly applicable to various TableQA and Table Fact Verification tasks.

Through extensive experiments across in-domain and out-of-domain datasets, we find that ATF significantly enhances generalization in challenging scenarios where tables are noisy, large, or structurally unfamiliar. In particular, models equipped with ATF show strong improvements in TableQA Out-of-domain datasets, demonstrating its potential to improve robustness under distribution shift.

Furthermore, ATF leads to substantial reductions in input length—up to 70\% fewer table cells—while still maintaining or improving downstream task performance. This cell-level compression is especially beneficial in practical settings where efficiency, interpretability, and latency are critical.

Overall, our findings highlight ATF as a promising preprocessing framework for table reasoning tasks, paving the way for future research on adaptive filtering, selective attention mechanisms, and integration with LLMs.

\section*{Limitations}

Despite its effectiveness, ATF presents several limitations. While it consistently improves performance on out-of-domain TableQA datasets, we observe performance degradation on in-domain TableQA and Table Fact Verification benchmarks. This indicates a potential distributional mismatch and contextual information loss, as models trained on unfiltered tables may struggle when evaluated on filtered ones during inference.

Moreover, ATF depends on multiple LLM queries for structural understanding and semantic scoring, introducing latency that can hinder real-time deployment, particularly in resource-constrained environments. The quality of filtering decisions is also affected by the domain coverage and consistency of the underlying LLM, which may vary across datasets.

Future work could address these issues by fine-tuning models on ATF-filtered tables to better align training and inference distributions, or by developing adaptive filtering strategies that selectively preserve context based on task-specific relevance.

\bibliographystyle{plainnat}
\bibliography{custom}

\begin{thebibliography}{33}
\providecommand{\natexlab}[1]{#1}
\providecommand{\url}[1]{\texttt{#1}}
\expandafter\ifx\csname urlstyle\endcsname\relax
  \providecommand{\doi}[1]{doi: #1}\else
  \providecommand{\doi}{doi: \begingroup \urlstyle{rm}\Url}\fi

\bibitem[Ben-David et~al.(2010)Ben-David, Blitzer, Crammer, and Pereira]{ben2010domain_adaptation}
Shai Ben-David, John Blitzer, Koby Crammer, and Fernando Pereira.
\newblock A theory of learning from different domains.
\newblock In \emph{Machine learning}, pages 79--103. Springer, 2010.

\bibitem[Chen et~al.(2024)Chen, Miculicich, Eisenschlos, Wang, Wang, Chen, Fujii, Lin, Lee, and Pfister]{chen2024tableragmilliontokentableunderstanding}
Si-An Chen, Lesly Miculicich, Julian~Martin Eisenschlos, Zifeng Wang, Zilong Wang, Yanfei Chen, Yasuhisa Fujii, Hsuan-Tien Lin, Chen-Yu Lee, and Tomas Pfister.
\newblock Tablerag: Million-token table understanding with language models, 2024.
\newblock URL \url{https://arxiv.org/abs/2410.04739}.

\bibitem[Chen et~al.(2020)Chen, Wang, Chen, Zhang, Wang, Li, Zhou, and Wang]{2019TabFactA}
Wenhu Chen, Hongmin Wang, Jianshu Chen, Yunkai Zhang, Hong Wang, Shiyang Li, Xiyou Zhou, and William~Yang Wang.
\newblock Tabfact: A large-scale dataset for table-based fact verification.
\newblock In \emph{Proceedings of the International Conference on Learning Representations (ICLR)}, Addis Ababa, Ethiopia, April 2020.

\bibitem[Cheng et~al.(2023)]{cheng2023binder}
Zhoujun Cheng et~al.
\newblock Binder: Binding language models in symbolic languages.
\newblock In \emph{Proceedings of the International Conference on Learning Representations (ICLR)}, 2023.

\bibitem[Devlin et~al.(2019)Devlin, Chang, Lee, and Toutanova]{devlin2019bertpretrainingdeepbidirectional}
Jacob Devlin, Ming-Wei Chang, Kenton Lee, and Kristina Toutanova.
\newblock Bert: Pre-training of deep bidirectional transformers for language understanding, 2019.
\newblock URL \url{https://arxiv.org/abs/1810.04805}.

\bibitem[Eisenschlos et~al.(2021)Eisenschlos, Duong, Ruder, Narayan, and Yogatama]{eisenschlos2021mate}
Julian Eisenschlos, Long~Phuoc Duong, Sebastian Ruder, Shashi Narayan, and Dani Yogatama.
\newblock Mate: Multi-view attention for table transformers.
\newblock In \emph{Findings of the Association for Computational Linguistics: EMNLP 2021}, 2021.

\bibitem[Farr et~al.(2024)Farr, Manzonelli, Cruickshank, Starbird, and West]{10826109}
David Farr, Nico Manzonelli, Iain Cruickshank, Kate Starbird, and Jevin West.
\newblock Llm chain ensembles for scalable and accurate data annotation.
\newblock In \emph{2024 IEEE International Conference on Big Data (BigData)}, pages 2110--2118, 2024.
\newblock \doi{10.1109/BigData62323.2024.10826109}.

\bibitem[Glenn et~al.(2024)]{glenn2024blendsql}
Parker Glenn et~al.
\newblock Blendsql: A scalable dialect for unifying hybrid qa in relational algebra, 2024.
\newblock Preprint.

\bibitem[Hendrycks and Gimpel(2020)]{hendrycks2020ood_detection}
Dan Hendrycks and Kevin Gimpel.
\newblock Anomaly detection using softmax probability of pre-trained deep neural networks.
\newblock \emph{arXiv preprint arXiv:1612.02685}, 2020.

\bibitem[Herzig et~al.(2020)Herzig, Nowak, M{\"u}ller, Piccinno, and Eisenschlos]{herzig2020tapas}
Jonathan Herzig, Pawe{\l} Nowak, Thomas M{\"u}ller, Francesco Piccinno, and Julian Eisenschlos.
\newblock Tapas: Weakly supervised table parsing via pre-training.
\newblock In \emph{Proceedings of the 58th Annual Meeting of the Association for Computational Linguistics (ACL)}, 2020.

\bibitem[James et~al.(2013)James, Witten, Hastie, and Tibshirani]{james2013introduction}
Gareth James, Daniela Witten, Trevor Hastie, and Robert Tibshirani.
\newblock \emph{An Introduction to Statistical Learning}.
\newblock Springer, 2013.

\bibitem[Jones(1972)]{jones1972statistical}
Karen~Spärck Jones.
\newblock A statistical interpretation of term specificity and its application in retrieval.
\newblock \emph{Journal of Documentation}, 28\penalty0 (1):\penalty0 11--21, 1972.

\bibitem[Katsis et~al.(2021)Katsis, Chemmengath, Kumar, Bharadwaj, Canim, Glass, Gliozzo, Pan, Sen, Sankaranarayanan, and Chakrabarti]{katsis2021aitqa}
Yannis Katsis, Saneem Chemmengath, Vishwajeet Kumar, Samarth Bharadwaj, Mustafa Canim, Michael Glass, Alfio Gliozzo, Feifei Pan, Jaydeep Sen, Karthik Sankaranarayanan, and Soumen Chakrabarti.
\newblock Ait-qa: Question answering dataset over complex tables in the airline industry, 2021.

\bibitem[Kweon et~al.(2023{\natexlab{a}})Kweon, Kwon, Cho, Jo, and Choi]{kweon2023openwikitabledatasetopendomain}
Sunjun Kweon, Yeonsu Kwon, Seonhee Cho, Yohan Jo, and Edward Choi.
\newblock Open-wikitable: Dataset for open domain question answering with complex reasoning over table, 2023{\natexlab{a}}.
\newblock URL \url{https://arxiv.org/abs/2305.07288}.

\bibitem[Kweon et~al.(2023{\natexlab{b}})]{kweon2023openwikitables}
Sunjun Kweon et~al.
\newblock Open-wikitable: Dataset for odqa with complex reasoning over table.
\newblock In \emph{ACL Findings}, 2023{\natexlab{b}}.

\bibitem[Lewis et~al.(2020)Lewis, Liu, Goyal, Ghazvininejad, Mohamed, Levy, Stoyanov, and Zettlemoyer]{lewis-etal-2020-bart}
Mike Lewis, Yinhan Liu, Naman Goyal, Marjan Ghazvininejad, Abdelrahman Mohamed, Omer Levy, Veselin Stoyanov, and Luke Zettlemoyer.
\newblock {BART}: Denoising sequence-to-sequence pre-training for natural language generation, translation, and comprehension.
\newblock In Dan Jurafsky, Joyce Chai, Natalie Schluter, and Joel Tetreault, editors, \emph{Proceedings of the 58th Annual Meeting of the Association for Computational Linguistics}, pages 7871--7880, Online, July 2020. Association for Computational Linguistics.
\newblock \doi{10.18653/v1/2020.acl-main.703}.
\newblock URL \url{https://aclanthology.org/2020.acl-main.703/}.

\bibitem[Liu et~al.(2022)Liu, Chen, Guo, Ziyadi, Lin, Chen, and Lou]{liu2022tapextablepretraininglearning}
Qian Liu, Bei Chen, Jiaqi Guo, Morteza Ziyadi, Zeqi Lin, Weizhu Chen, and Jian-Guang Lou.
\newblock Tapex: Table pre-training via learning a neural sql executor, 2022.
\newblock URL \url{https://arxiv.org/abs/2107.07653}.

\bibitem[Nguyen et~al.(2025)Nguyen, Brugere, Sharma, Kariyappa, Nguyen, and Lecue]{nguyen2025interpretablellmbasedtablequestion}
Giang Nguyen, Ivan Brugere, Shubham Sharma, Sanjay Kariyappa, Anh~Totti Nguyen, and Freddy Lecue.
\newblock Interpretable llm-based table question answering, 2025.
\newblock URL \url{https://arxiv.org/abs/2412.12386}.

\bibitem[OpenAI(2024)]{openai2024gpt4o}
OpenAI.
\newblock Gpt-4o: Openai’s new multimodal model.
\newblock \url{https://openai.com/index/gpt-4o}, May 2024.
\newblock Accessed: 2025-06-02.

\bibitem[Pasupat and Liang(2015)]{pasupat2015compositionalsemanticparsingsemistructured}
Panupong Pasupat and Percy Liang.
\newblock Compositional semantic parsing on semi-structured tables, 2015.
\newblock URL \url{https://arxiv.org/abs/1508.00305}.

\bibitem[Reimers and Gurevych(2019)]{reimers2019sbert}
Nils Reimers and Iryna Gurevych.
\newblock Sentence-bert: Sentence embeddings using siamese bert-networks.
\newblock In \emph{Proceedings of the 2019 Conference on Empirical Methods in Natural Language Processing (EMNLP)}, pages 3982--3992, 2019.
\newblock URL \url{https://aclanthology.org/D19-1410}.

\bibitem[Robertson and Walker(1994)]{robertson1994bm25}
Stephen Robertson and Steve Walker.
\newblock Some simple effective approximations to the 2-poisson model for probabilistic weighted retrieval.
\newblock \emph{SIGIR}, pages 232--241, 1994.

\bibitem[Rousseeuw(1987)]{rousseeuw1987silhouettes}
Peter~J Rousseeuw.
\newblock Silhouettes: a graphical aid to the interpretation and validation of cluster analysis.
\newblock \emph{Journal of computational and applied mathematics}, 20:\penalty0 53--65, 1987.

\bibitem[Seo et~al.(2024)Seo, Yeo, and Lee]{seo2024unveilingimplicittableknowledge}
Kwangwook Seo, Jinyoung Yeo, and Dongha Lee.
\newblock Unveiling implicit table knowledge with question-then-pinpoint reasoner for insightful table summarization, 2024.
\newblock URL \url{https://arxiv.org/abs/2406.12269}.

\bibitem[Vaswani et~al.(2017)Vaswani, Shazeer, Parmar, Uszkoreit, Jones, Gomez, Kaiser, and Polosukhin]{vaswani2017attention}
Ashish Vaswani, Noam Shazeer, Niki Parmar, Jakob Uszkoreit, Llion Jones, Aidan~N Gomez, {\L}ukasz Kaiser, and Illia Polosukhin.
\newblock Attention is all you need.
\newblock In \emph{Advances in neural information processing systems}, pages 5998--6008, 2017.

\bibitem[Wang et~al.(2022)Wang, Wei, Schuurmans, Le, Chi, Zhou, et~al.]{wang2022self}
Jason Wang, Jason Wei, Dale Schuurmans, Quoc~V Le, Ed~H Chi, Denny Zhou, et~al.
\newblock Self-consistency improves chain of thought reasoning in language models.
\newblock \emph{arXiv preprint arXiv:2203.11171}, 2022.

\bibitem[Wang et~al.(2021)Wang, Mahajan, Danilevsky, and Rosenthal]{wang-etal-2021-semeval}
Nancy X.~R. Wang, Diwakar Mahajan, Marina Danilevsky, and Sara Rosenthal.
\newblock {S}em{E}val-2021 task 9: Fact verification and evidence finding for tabular data in scientific documents ({SEM}-{TAB}-{FACTS}).
\newblock In Alexis Palmer, Nathan Schneider, Natalie Schluter, Guy Emerson, Aurelie Herbelot, and Xiaodan Zhu, editors, \emph{Proceedings of the 15th International Workshop on Semantic Evaluation (SemEval-2021)}, pages 317--326, Online, August 2021. Association for Computational Linguistics.
\newblock \doi{10.18653/v1/2021.semeval-1.39}.
\newblock URL \url{https://aclanthology.org/2021.semeval-1.39/}.

\bibitem[Wang et~al.(2024)Wang, Zhang, Li, Eisenschlos, Perot, Wang, Miculicich, Fujii, Shang, Lee, and Pfister]{wang2024chainoftableevolvingtablesreasoning}
Zilong Wang, Hao Zhang, Chun-Liang Li, Julian~Martin Eisenschlos, Vincent Perot, Zifeng Wang, Lesly Miculicich, Yasuhisa Fujii, Jingbo Shang, Chen-Yu Lee, and Tomas Pfister.
\newblock Chain-of-table: Evolving tables in the reasoning chain for table understanding, 2024.
\newblock URL \url{https://arxiv.org/abs/2401.04398}.

\bibitem[Wu and Feng(2024)]{wu2024protrix}
Zirui Wu and Yansong Feng.
\newblock Protrix: Planning and reasoning over tables with sentence context.
\newblock In \emph{Proceedings of the 2024 Conference on Empirical Methods in Natural Language Processing (EMNLP)}, 2024.

\bibitem[Ye et~al.(2023)Ye, Hui, Yang, Li, Huang, and Li]{ye2023largelanguagemodelsversatile}
Yunhu Ye, Binyuan Hui, Min Yang, Binhua Li, Fei Huang, and Yongbin Li.
\newblock Large language models are versatile decomposers: Decompose evidence and questions for table-based reasoning, 2023.
\newblock URL \url{https://arxiv.org/abs/2301.13808}.

\bibitem[Zhang et~al.(2025)Zhang, Ma, and Yang]{zhang-etal-2025-alter}
Han Zhang, Yuheng Ma, and Hanfang Yang.
\newblock {ALTER}: Augmentation for large-table-based reasoning.
\newblock In Luis Chiruzzo, Alan Ritter, and Lu~Wang, editors, \emph{Proceedings of the 2025 Conference of the Nations of the Americas Chapter of the Association for Computational Linguistics: Human Language Technologies (Volume 1: Long Papers)}, pages 179--198, Albuquerque, New Mexico, April 2025. Association for Computational Linguistics.
\newblock ISBN 979-8-89176-189-6.
\newblock \doi{10.18653/v1/2025.naacl-long.9}.
\newblock URL \url{https://aclanthology.org/2025.naacl-long.9/}.

\bibitem[Zhang et~al.(2023)Zhang, Henkel, Floratou, Cahoon, Deep, and Patel]{zhang2023reactableenhancingreacttable}
Yunjia Zhang, Jordan Henkel, Avrilia Floratou, Joyce Cahoon, Shaleen Deep, and Jignesh~M. Patel.
\newblock Reactable: Enhancing react for table question answering, 2023.
\newblock URL \url{https://arxiv.org/abs/2310.00815}.

\bibitem[Zhu et~al.(2021)Zhu, Lei, Huang, Wang, Zhang, Lv, Feng, and Chua]{zhu2021tatqaquestionansweringbenchmark}
Fengbin Zhu, Wenqiang Lei, Youcheng Huang, Chao Wang, Shuo Zhang, Jiancheng Lv, Fuli Feng, and Tat-Seng Chua.
\newblock Tat-qa: A question answering benchmark on a hybrid of tabular and textual content in finance, 2021.
\newblock URL \url{https://arxiv.org/abs/2105.07624}.

\end{thebibliography}

\appendix
\section{Cluster Selection Details}
\label{appendix:cluster_selection}

In this section, we provide technical details for the three selection strategies introduced in Section~3.5: \textit{semantic similarity, multi-criteria decision making (MCDM), and adaptive confidence-based thresholding}. These methods operate on column-level score pairs $\{\hat{s}^{\text{final}}_i, s^{\text{emb}}_i\}$ and guide the cluster-level selection process and we finally select optimal cluster by voting for each cluster.

\subsection{Semantic Similarity with Cluster Quality}

\textit{Semantic Similarity with Cluster Quality} is focused to similarity of question and cluster centroid of each column cluster for this process we use cluster quality weight (intra-cluster cohesion and inter-cluster separation) for accuracy and stability of cluster selection.

Let $v_q$ be the question vector and $v_{C_j}$ be the centroid of cluster $C_j$. The selection score is given by:
\begin{equation}
\text{Score}(C_j) = \text{sim}(v_q, v_{C_j}) \cdot Q(C_j)
\end{equation}
where $Q(C_j)$ is a cluster quality weight computed as:
\begin{equation}
Q(C_j) = 0.6 \cdot \text{Cohesion}(C_j) + 0.4 \cdot \text{Separation}(C_j)
\end{equation}
We select the cluster with the highest score.

\subsection{Multi-Criteria Decision Making (MCDM)}

\textit{Multi-Criteria Decision Making (MCDM)} is focused to evaluate each cluster based on various criteria:

\begin{itemize}
  \item \textbf{Relevance Score ($R_j$)}: Average relevance score across all columns in cluster $C_j$. Each column's relevance is computed from its combined LLM and embedding-based score vector.
  \begin{equation}
  R_j = \frac{1}{|C_j|} \sum_{c_i \in C_j} \text{mean}(s_i)
  \end{equation}

  \item \textbf{Diversity Score ($D_j$)}: Lexical diversity based on column name components. We extract prefixes and suffixes of column names to estimate diversity:
  \begin{equation}
  D_j = \frac{|\text{prefixes}(C_j)| + |\text{suffixes}(C_j)|}{2 \cdot |C_j| + \varepsilon}
  \end{equation}
  where $\varepsilon$ is a small constant for numerical stability.

  \item \textbf{Information Density ($I_j$)}: Ratio of columns with average score above a fixed threshold $\tau$ (e.g., $\tau = 0.7$):
  \begin{equation}
  I_j = \frac{1}{|C_j|} \sum_{c_i \in C_j} \mathbb{I}[\text{mean}(s_i) > \tau]
  \end{equation}
  where $\mathbb{I}[\cdot]$ is the indicator function.

  \item \textbf{Size-Complexity Match ($M_j$)}: Measures how well the number of columns in the cluster aligns with the question’s estimated complexity score $Q_c$. A soft penalty is applied for size deviation:
  \begin{equation}
  M_j = \frac{1}{1 + \alpha \cdot |\,|C_j| - \text{optimal}(Q_c)\,|}
  \end{equation}
  where $\alpha$ is a scaling factor and $\text{optimal}(Q_c) = \min(3Q_c, 10)$ controls ideal cluster size based on complexity.
\end{itemize}

\vspace{0.5em}
The final score is a weighted sum:
\begin{equation}
\text{Score}(C_j) = w_1 \cdot R_j + w_2 \cdot D_j + w_3 \cdot I_j + w_4 \cdot M_j
\end{equation}
where $(w_1, w_2, w_3, w_4) = (0.4, 0.2, 0.2, 0.2)$.

This strategy enables selection of clusters that are not only relevant but also diverse, dense in useful information, and appropriately sized for the given question context.

\subsection{Adaptive Thresholding and Confidence Scoring}
\label{appendix:adaptive_thresholding}

\textit{Adaptive Thresholding and Confidence Scoring} is designed to dynamically select the most reliable column cluster by quantifying and comparing confidence levels across clusters.

We define the confidence score for each cluster $C_j$ as a weighted combination of three factors:

\begin{equation}
\text{Conf}(C_j) = 0.4 \cdot \text{Cons}_j + 0.4 \cdot \text{Stren}_j + 0.2 \cdot \text{Type}_j
\end{equation}

\begin{itemize}
    \item \textbf{Cons}$_j$: Measures the homogeneity of column-level scores within the cluster, computed as the inverse of score variance.
    \item \textbf{Stren}$_j$: Represents the average score magnitude of columns in the cluster.
    \item \textbf{Type}$_j$: A prior weight derived from the question type (e.g., aggregation, filtering, exploration) and the cluster size, based on task-specific preferences.
\end{itemize}

Once all confidence scores $\{\text{Conf}(C_j)\}$ are computed, a dynamic threshold $\tau$ is set to filter low-quality clusters:

\begin{equation}
\tau = \mu - 0.5\sigma
\end{equation}

where $\mu$ and $\sigma$ denote the mean and standard deviation of the confidence scores across all clusters.

Clusters whose confidence exceeds $\tau$ are considered valid candidates. Among these, the cluster with the highest confidence is selected. If no clusters meet the threshold, the most confident cluster is selected regardless of thresholding.

\subsection{Ensemble Voting Strategy}
\label{appendix:ensemble_voting}

We combine the outputs of the three independent cluster selection strategies—semantic similarity, multi-criteria decision making (MCDM), and adaptive confidence scoring—through an ensemble voting mechanism.

Let $\{C_1, C_2, C_3\}$ denote the cluster selections by each method. The final cluster $C^*$ is chosen as the mode (majority vote) of the three:

\begin{equation}
C^* = \text{mode}(C_1, C_2, C_3)
\end{equation}

In cases of a tie (i.e., all three strategies return different clusters), we break the tie by selecting the cluster associated with the highest confidence score among the three methods.

After determining the primary cluster $C^*$, we retain all columns within it. To ensure diverse information coverage, we also include the top-$k$ most relevant columns from each non-selected cluster, where $k$ is a small constant (typically $k=1$). This guarantees that useful but otherwise underrepresented columns are not excluded.

Furthermore, essential columns—such as known key identifiers or answer-bearing columns—are always preserved, even if they do not belong to the selected cluster set.

This ensemble approach enhances robustness by aggregating multiple perspectives, balances global relevance with local diversity, and ensures that important information is not lost due to the limitations of any single strategy.

\vspace{0.3em}
\noindent\textbf{Time Complexity}
\label{appendix:complexity}
In this cluster selection algorithm, some steps may be computationally expensive depending on the table size and number of clusters. To analyze the cost, we compute the time complexity of each component involved in the pipeline.

We summarize the time complexity of each selection strategy component in Table~\ref{tab:simple_complexity}. Here, $m$ is the number of columns, $k$ is the number of clusters, $d$ is the embedding dimension, and $t$ is the number of K-Means iterations.

\begin{table}[htbp]
\centering
\begin{tabular}{l|l}
\toprule
\textbf{Component} & \textbf{Time Complexity} \\
\midrule
K-means Clustering & $\mathcal{O}(m \cdot k \cdot d)$ \\
MCDM Strategy & $\mathcal{O}(m)$ \\
Adaptive Thresholding & $\mathcal{O}(m)$ \\
Ensemble Voting & $\mathcal{O}(k)$ \\
\bottomrule
\end{tabular}
\caption{Time complexity of each cluster selection component.}
\label{tab:simple_complexity}
\end{table}

\noindent\textbf{Comment.} Final ensemble step is computationally negligible and ensures robustness.

\vspace{0.5em}
\noindent Overall, all methods are designed to operate in linear or near-linear time with respect to the number of columns $m$, making the framework scalable for large real-world tables.

In case of a tie, the method-specific confidence scores (or selection scores) are compared, and the cluster with the highest value is selected. To ensure diversity, we also include top-$k$ columns from non-selected clusters, ranked by relevance.

\section{Implementation Details}

Our proposed ATF framework is implemented using Python 3.10 with PyTorch 2.1 and HuggingFace Transformers. All experiments are conducted on a single NVIDIA A100 GPU (83.5GB) via Google Colab Pro unless otherwise noted.

The column and row filtering steps of ATF were performed on a MacBook Pro M3 (14-core GPU), which was used to generate the filtered table datasets prior to evaluation and experiments.

\begin{itemize}
    \item \textbf{LLM Components:} \\
    We utilize the \texttt{gpt-4-mini} API for multiple reasoning steps in the ATF pipeline, including: \textit{(i)} answer entity type prediction, \textit{(ii)} essential column extraction, \textit{(iii)} column description generation, and \textit{(iv)} column scoring.  
    All LLM calls are performed with \texttt{temperature = 0.0} to ensure deterministic outputs.

    \item \textbf{Dense Similarity Encoder:} \\
    For semantic similarity computation in both column- and row-level relevance scoring, we use the \texttt{all-MiniLM-L6-v2} model from HuggingFace.
\end{itemize}

\section{ATF Algorithms}

Algorithm~\ref{alg:column_filtering} outlines the full procedure of column-level filtering in the ATF framework.
Algorithm~\ref{alg:row_filtering} presents the ATF row-level filtering pipeline, which performs hybrid scoring over the selected columns and constructs a compact final table representation.

\begin{algorithm*}[t]
\caption[ATF Column Filtering]{ATF Column-Level Filtering Algorithm. This algorithm describes the multi-step selection of columns.}
\label{alg:column_filtering}
\begin{algorithmic}[1]
\Require Question $q$, headers $\mathcal{H} = \{h_1, ..., h_m\}$, cell values $V_i$
\Ensure Filtered column set $C'$
\State $e^* \gets \mathsf{LLM}_{\mathsf{CoT}}(q, \mathcal{E})$ \Comment{Predict answer entity type}
\State $C_{\text{essential}} \gets \mathsf{LLM}(q, \mathcal{H})$ \Comment{Extract essential columns}
\ForAll{$h_i \in \mathcal{H}$}
    \State $d_i \gets \mathsf{LLM}_{\mathsf{desc}}(h_i, V_i)$
\EndFor
\ForAll{$h_i \in \mathcal{H}$}
    \For{$t = 1$ to $N$}
        \State $s_i^{(t)} \gets \mathsf{LLM}_{\mathsf{score}}^{(t)}(q, d_i, e^*, V_i)$
    \EndFor
    \State $\mu_i \gets \frac{1}{N} \sum_{t=1}^{N} s_i^{(t)}$
    \State $\sigma_i \gets \sqrt{\frac{1}{N} \sum_{t=1}^N (s_i^{(t)} - \mu_i)^2}$
    \State $\hat{s}_i^{\mathsf{col\text{-}final}} \gets \mu_i \cdot \left(\frac{1}{1 + \sigma_i}\right)$
    \State $s_i^{\mathsf{emb}} \gets \mathsf{cos\_sim}(\mathsf{emb}(q),\ \mathsf{emb}(h_i \oplus d_i))$
\EndFor
\State $\mathcal{C} \gets \mathsf{KMeans}(\{[\hat{s}_i^{\mathsf{col\text{-}final}},\ s_i^{\mathsf{emb}}]\}_{i=1}^m)$
\ForAll{cluster $C_j \in \mathcal{C}$}
    \State Compute:
    \Statex \quad (1) semantic similarity: $\mathsf{sim}(v_q, v_{C_j}) \cdot Q(C_j)$
    \Statex \quad (2) multi-criteria: $w_1 R_j + w_2 D_j + w_3 I_j + w_4 M_j$
    \Statex \quad (3) confidence: $0.4 \cdot \mathsf{Cons}_j + 0.4 \cdot \mathsf{Stren}_j + 0.2 \cdot \mathsf{Type}_j$
\EndFor
\State $C^* \gets \mathsf{mode}(C_1, C_2, C_3)$
\State $C' \gets C_{C^*} \cup \bigcup_{C_j \ne C^*} \mathsf{Top}_1(C_j)$
\State $C' \gets C' \cup C_{\text{essential}}$
\Return $C'$
\end{algorithmic}
\end{algorithm*}

\vspace{1em}

\begin{algorithm*}[t]
\caption{ATF Row-Level Filtering and Final Table Construction}
\label{alg:row_filtering}
\begin{algorithmic}[1]
\Require Question $q$, filtered columns $C'$, table $T$ with $n$ rows and $m$ columns
\Ensure Compressed table $T'$
\For{$i = 1$ to $n$}
    \State $r_i^{\text{text}} \gets \bigoplus_{c_j \in C'} T[r_i, c_j]$
\EndFor
\For{$i = 1$ to $n$}
    \State $s_i^{\text{TF-IDF}} \gets \text{cosine\_sim}(\text{tfidf}(q), \text{tfidf}(r_i^{\text{text}}))$
    \State $s_i^{\text{BM25}} \gets \text{BM25}(q, r_i^{\text{text}})$
    \State $s_i^{\text{dense}} \gets \text{cosine\_sim}(\text{emb}(q), \text{emb}(r_i^{\text{text}}))$
\EndFor
\State Apply softmax normalization:
\For{$i = 1$ to $n$}
    \State $\tilde{s}_i^{\text{TF-IDF}} \gets \text{softmax}(s_i^{\text{TF-IDF}})$
    \State $\tilde{s}_i^{\text{BM25}} \gets \text{softmax}(s_i^{\text{BM25}})$
    \State $\tilde{s}_i^{\text{dense}} \gets \text{softmax}(s_i^{\text{dense}})$
\EndFor
\For{$i = 1$ to $n$}
    \State $s_i^{\text{row-final}} \gets 0.4 \cdot \tilde{s}_i^{\text{TF-IDF}} + 0.3 \cdot \tilde{s}_i^{\text{BM25}} + 0.3 \cdot \tilde{s}_i^{\text{dense}}$
\EndFor
\State $k \gets \left\lceil \alpha \cdot n \right\rceil$, where $\alpha = 0.4$
\State $R' \gets$ indices of top-$k$ rows based on $s_i^{\text{row-final}}$
\State $T' \gets T[R', C']$
\Return $T'$
\end{algorithmic}
\end{algorithm*}

\section{ATF LLM Prompts}

To enhance table understanding and column selection in the ATF pipeline, we employed several LLM prompts at different reasoning stages. Below, we summarize the purpose of each prompt.

\begin{itemize}
    \item \textbf{Answer Entity Type Prediction Prompt} (Figure~\ref{fig:entity_prompt}): Infers the expected answer type (e.g., number, organization) to guide downstream filtering and relevance scoring.
    
    \item \textbf{Essential Column Selection Prompt} (Figure~\ref{fig:essential_column_prompt}): Selects a minimal subset of columns from the raw table deemed essential for answering the question.

    \item \textbf{Column Description Generation Prompt} (Figure~\ref{fig:col_description_prompt}): Produces natural language descriptions of each column to enable semantic scoring and LLM-based ranking.

    \item \textbf{Column Relevance Scoring Prompt} (Figure~\ref{fig:col_scoring_prompt}): Assigns a 0.0–1.0 relevance score to each column, using both question context and generated descriptions.
\end{itemize}

\begin{figure*}[t]
\centering
\caption{\textbf{Prompt used for Answer Entity Type Prediction by LLM}}
\label{fig:entity_prompt}
\begin{minipage}{0.95\textwidth}
\small\raggedright
\noindent
\texttt{You are a Question Answering expert.}

\medskip

\texttt{First, carefully read the question and rephrase it in a clearer and more specific way that makes the target of the answer obvious.}

\medskip

\texttt{Then, based on the rephrased question, determine what type of entity the answer is asking for. This might be:} \\
\texttt{- a person (e.g., someone's name),} \\
\texttt{- an organization (e.g., a team, company, etc.),} \\
\texttt{- a date (e.g., a specific day),} \\
\texttt{- a number (e.g., age, count, price),} \\
\texttt{- a location (e.g., country, city, place),} \\
\texttt{- or some other type.}

\medskip

\texttt{You may include new types if needed. Then, assign a confidence score (0.0 to 1.0) to each type, depending on how likely it is to be the expected answer.}

\medskip

\texttt{Respond in exactly this JSON format:} \\
\texttt{\{} \\
\texttt{\ \ \ "EntityType1": score,} \\
\texttt{\ \ \ "EntityType2": score} \\
\texttt{\}}

\medskip

\texttt{Question: \{question\}}
\end{minipage}
\end{figure*}

\begin{figure*}[t]
\centering
\caption{\textbf{Prompt used for Extracting Essential Columns from the Table}}
\label{fig:essential_column_prompt}
\begin{minipage}{0.95\textwidth}
\small\raggedright
\noindent
\texttt{You are a table question answering expert.}

\medskip

\texttt{Your task is to identify the essential table columns required to answer a given question, from the list of available columns.}

\medskip

\texttt{Please strictly follow these instructions:} \\
\texttt{- Output only a Python list of column names (e.g., ["ColumnA", "ColumnB"])} \\
\texttt{- Do not include any explanation or extra text.} \\
\texttt{- If you're unsure, choose conservatively by selecting columns that appear most relevant to keywords in the question.} \\
\texttt{- Use exact column names as they appear in the list.}

\medskip

\texttt{Question: \{question\}}

\texttt{Available Columns: \{Column1, Column2, ..., ColumnN\}}

\medskip

\texttt{Expected Answer Type: \{predicted\_answer\_entity\} (Optional)}

\medskip

\texttt{Essential Columns:}
\end{minipage}
\end{figure*}

\begin{figure*}[t]
\centering
\caption{\textbf{Prompt used for Column Description Generation by LLM}}
\label{fig:col_description_prompt}
\begin{minipage}{0.95\textwidth}
\small\raggedright
\noindent
\texttt{You are a table analysis expert. Generate concise, consistent descriptions for each column.}

\medskip

\texttt{Question Context: \{question\}}

\texttt{Expected Answer Type: \{predicted\_entity\} (Optional)}

\medskip

\texttt{Columns with Examples:} \\
\texttt{Column1 (Examples: ...)} \\
\texttt{Column2 (Examples: ...)} \\
\texttt{...}

\medskip

\texttt{Rules:} \\
\texttt{1. Keep descriptions factual and concise (max 15 words)} \\
\texttt{2. Focus on what the column represents, not just examples} \\
\texttt{3. Use consistent terminology} \\
\texttt{4. Include data type when relevant}

\medskip

\texttt{Format:} \\
\texttt{Column1: description} \\
\texttt{Column2: description}
\end{minipage}
\end{figure*}

\begin{figure*}[t]
\centering
\caption{\textbf{Prompt used for Column Relevance Scoring by LLM}}
\label{fig:col_scoring_prompt}
\begin{minipage}{0.95\textwidth}
\small\raggedright
\noindent
\texttt{You are an expert in question-answering with tabular data.}

\medskip

\texttt{Question: \{question\}} \\
\texttt{Expected Answer Type: \{predicted\_entity\} (Optional)}

\medskip

\texttt{Table Columns:} \\
\texttt{Column1: description} \\
\texttt{Column2: description} \\
\texttt{...}

\medskip

\texttt{Rate each column's relevance for answering the question (0.0 to 1.0):} \\
\texttt{- 1.0: Essential/Primary key for the answer} \\
\texttt{- 0.8: Highly relevant, likely needed} \\
\texttt{- 0.6: Moderately relevant, could be useful} \\
\texttt{- 0.4: Somewhat relevant, might provide context} \\
\texttt{- 0.2: Low relevance, probably not needed} \\
\texttt{- 0.0: Not relevant at all}

\medskip

\texttt{Be consistent and precise. Consider:} \\
\texttt{1. Direct relevance to the question} \\
\texttt{2. Potential for filtering/grouping} \\
\texttt{3. Contextual importance}

\medskip

\texttt{Format (exact format required):} \\
\texttt{column\_name: score}
\end{minipage}
\end{figure*}

\section{ATF Case Study}

We select representative examples of ATF and
present in Table 9 to 12


\begin{table*}[t]
\centering
\small
\caption{Example of the ATF filtering process on an Open-WikiTable instance (index = 777)}
\label{tab:ATF_example of openwiki}
\begin{tabular}{p{0.22\linewidth} | p{0.72\linewidth}}
\toprule
\multicolumn{2}{c}{\textbf{Example filtering process of ATF from Open-WikiTable}} \\
\midrule
\textbf{Question (Input)} & \textit{“What money (\$) has a score of 70-66-73-69=278 at the 1993 US Open (golf) finals?”} \\

\midrule
\textbf{Raw Table (Input)} & 
\begin{tabular}[t]{@{}l@{}}
col: \texttt{Place | Player | Country | Score | To\_par | Money\_} \\
row 0: 1 | Lee Janzen | United States | 67-67-69-69=272 | –8 | 290000.0 \\ 
row 1: 2 | Payne Stewart | United States | 70-66-68-70=274 | –6 | 145000.0 \\
row 2: T3 | Paul Azinger | United States | 71-68-69-69=277 | –3 | 78556.0 \\
row 3: T3 | Craig Parry | Australia | 66-74-69-68=277 | –3 | 78556.0 \\ 
... \\
row 6: T7 | Ernie Els | South Africa | 71-73-68-67=279 | –1 | 35481.0 \\
row 7: T7 | Raymond Floyd | United States | 68-73-70-68=279 | –1 | 35481.0 \\
row 8: T7 | Fred Funk | United States | 70-72-67-70=279 | –1 | 35481.0 \\
row 9: T7 | Nolan Henke | United States | 72-71-67-69=279 | –1 | 35481.0 \\ 
\end{tabular}\\

\midrule
\textbf{Predicted Anser Entity} & \textit{number} \\

\midrule
\textbf{Column Description} & 
\begin{tabular}[t]{@{}l@{}}
\texttt{Place}: Final standing of the player in the tournament (e.g., T7, T3, T5, 1, 2) \\ 
\texttt{Player}: Name of the golfer participating in the event (e.g., Lee Janzen, Payne Stewart, ...) \\
\texttt{Country}: Nationality of the player (e.g., United States, Australia, South Africa) \\
\texttt{Score}: Total strokes taken over the tournament (e.g., 67-67-69-69=272, ...) \\
\texttt{To\_par}: Total score relative to par (e.g., –1, –3, –2, –8, –6) \\
\texttt{Money\_}: Prize money earned by the player  (e.g., 35481, 78556, ...) \\
\end{tabular} \\

\midrule
\textbf{Column Relevance scores} & 
\begin{tabular}[t]{@{}l@{}}
\textbf{\textit{Score}: [LLM Score, Cosine Score]} \\
Place: [0.0, 0.2674] \\ 
Player: [0.0, 0.4826] \\ 
Country: [0.0, 0.0] \\ 
Score: [1.0, 1.0] \\ 
To\_par: [0.0, 0.507] \\ 
Money\_: [1.0, 0.9339] \\
\end{tabular}
\\

\midrule
\textbf{Column Clustering} & 
\begin{tabular}[t]{@{}l@{}}
\textit{Cluster 0}: [\texttt{Country}] → Select \texttt{Country} \\
\textit{Cluster 1}: [\texttt{Score}, \texttt{Money\_}] → Selected Cluster → Select ALL \\
\textit{Cluster 2}: [\texttt{Place}, \texttt{Player}, \texttt{To\_par}] → Select \texttt{To\_par} \\
\end{tabular}
\\

\midrule
\textbf{Essential Columns} & 
Essential Columns: [\texttt{Score}, \texttt{Money\_}]\\

\midrule
\textbf{Column Selection} & 
Essentail Columns + Cluster Selected Columns \newline
\textbf{Final Selected Columns: [\texttt{Score}, \texttt{Money\_}, \texttt{Country}, \texttt{To\_par}]} \\

\midrule
\textbf{Flatten Row text} & 
\begin{tabular}[t]{@{}l@{}}
col: \texttt{Country Score To\_par Money\_} \\
row 0: United States 67-67-69-69=272 –8 290000.0 \\ 
row 1: United States 70-66-68-70=274 –6 145000.0 \\
... \\
row 8: United States 70-72-67-70=279 –1 | 35481.0 \\
row 9: United States 72-71-67-69=279 –1 | 35481.0 \\ 
\end{tabular}
\\

\midrule
\textbf{Row Hybrid Scores} & 
\begin{tabular}[t]{@{}l@{}}
\textbf{\textit{Score}: [TF-IDF Score, BM25 Score, Dense Score]} \\
row 0: [0.0695, 0.0, 0.4418] \\ 
row 1: [0.1377, 0.0, 0.4635] \\
... \\
row 8: [0.1004, 0.0, 0.3988] \\
row 9: [0.0452, 0.0, 0.3556] \\ 
\texttt{----------------------------------------------------------------------------------------------------------------------------------------------------------------------------------------------------------------------------} \\
Final Score: 0.4 $\cdot$ softmax(TF-IDF) + 0.3 $\cdot$ softmax(BM25) + 0.3 $\cdot$ softmax(Dense)
\end{tabular}
\\

\midrule
\textbf{Row Selection} & 
Select Top-40\% → \textbf{Final Selected Row indices: [5.0, 1.0, 4.0, 7.0]}
\newline
(Selected row final scores: [0.2025, 0.0926, 0.0899, 0.0893]) \\

\midrule
\textbf{Filtered Table (Output)} & 
\begin{tabular}[t]{@{}l@{}}
col: \texttt{Country | Score | To\_par | Money\_} \\
row 5: United States | 70-66-73-69=278 | –2 | \textbf{48730.0} \\
row 1: United States | 70-66-68-70=274 | –6 | 145000.0 \\
row 4: United States | 66-72-72-68=278 | –2 | 48730.0 \\ 
row 7: United States | 68-73-70-68=279 | –1 | 35481.0 \\
\end{tabular}\\

\midrule
\textbf{Real Answer} & \textbf{\texttt{['48730.0']}} \\
\bottomrule
\end{tabular}
\end{table*}


\begin{table*}[t]
\centering
\small
\caption{Example of the ATF filtering process on a TabFact instance (7th statement from the 77th table)}
\label{tab:ATF_example of tabfact}
\begin{tabular}{p{0.22\linewidth} | p{0.72\linewidth}}
\toprule
\multicolumn{2}{c}{\textbf{Example filtering process of ATF from TabFact}} \\
\midrule
\textbf{Statement (Input)} & \textit{“kristofer martin have be nominate for 7 best actor award between 2012 and 2013 , win 4 of those award”} \\

\midrule
\textbf{Raw Table (Input)} & 
\begin{tabular}[t]{@{}l@{}}
col: \texttt{year | award giving body | category | nominated for | result} \\
row 0: 2013 | 10th golden screen awards | best actor in a supporting role | oros | won \\ 

row 1: 2012 | the young critics circle awards  | best performance | oros | nominated \\

row 2: 2012 | 28th pmpc ... for movies | new ... the year | tween ... of 2012 | nominated \\

row 3: 2012 | famas awards | german moreno youth achievement award | n / a | won \\ 

row 4: 2012 | the young critics circle awards  | best performance | oros | nominated \\

row 5: 2012 | 28th pmpc ... for movies | new ... the year | tween ... of 2012 | nominated \\

row 6: 2012 | famas awards | german moreno youth achievement award | n / a | won \\ 
\end{tabular}\\

\midrule
\textbf{Predicted Anser Entity} & \textit{person} \\

\midrule
\textbf{Column Description} & 
\begin{tabular}{@{}p{\linewidth}@{}}
\texttt{year}: The year the award was presented (e.g., 2012, 2013) \\ 
\texttt{award giving body}: The organization or event presenting the award (e.g., 1st sunday all stars awards, ...) \\
\texttt{category}: The specific award category ... nomination is made (e.g., stand out season performer, ...) \\
\texttt{nominated for}: The title or project associated with the nomination (e.g., oros, n / a, ...) \\
\texttt{result}: The outcome of the nomination, indicating if the award was won or not (e.g., won, nominated) \\
\end{tabular} \\

\midrule
\textbf{Column Relevance scores} & 
\begin{tabular}[t]{@{}l@{}}
\textbf{\textit{Score}: [LLM Score, Cosine Score]} \\
year: [0.792, 0.0498] \\ 
award giving body: [0.6092, 0.1648] \\ 
category: [0.4265, 1.0] \\ 
nominated for: [0.4, 0.2543] \\ 
result: [1.0, 0.0] \\ 
\end{tabular}
\\

\midrule
\textbf{Column Clustering} & 
\begin{tabular}[t]{@{}l@{}}
\textit{Cluster 0}: [\texttt{award giving body}, \texttt{nominated for}] → Select \texttt{award giving body} \\
\textit{Cluster 1}: [\texttt{category}] → Selected Cluster → Select ALL \\
\textit{Cluster 2}: [\texttt{year}, \texttt{result}] → Select \texttt{result} \\
\end{tabular}
\\

\midrule
\textbf{Essential Columns} & 
Essential Columns: [\texttt{year}, \texttt{nominated for}, \texttt{result}] \\

\midrule
\textbf{Column Selection} & 
Essential Columns + Cluster Selected Columns \newline
\textbf{Final Selected Columns: [\texttt{category}, \texttt{award giving body}, \texttt{result}, \texttt{year}, \texttt{nominated for}]} \\

\midrule
\textbf{Flatten Row text} & 
\begin{tabular}[t]{@{}l@{}}
col: \texttt{year award giving body category nominated for result} \\
row 0: 2013 10th golden screen awards best actor in a supporting role oros won \\ 
row 1: 2012 the young critics circle awards best performance oros nominated \\
row 2: 2012 28th pmpc ... for movies new ... the year tween ... of 2012 nominated \\
row 3: 2012 famas awards german moreno youth achievement award n / a won \\ 
row 4: 2012 the young critics circle awards best performance oros nominated \\
row 5: 2012 28th pmpc ... for movies new ... the year tween ... of 2012 nominated \\
row 6: 2012 famas awards german moreno youth achievement award n / a won \\  
\end{tabular}
\\

\midrule
\textbf{Row Hybrid Scores} & 
\begin{tabular}{@{}p{\linewidth}@{}}
\textbf{\textit{Score}: [TF-IDF Score, BM25 Score, Dense Score]} \\
row 0: [0.0335, 0.2697, 0.5631] \\ 
row 1: [0.1345, 1.5015, 0.4084] \\
... \\
row 5: [0.1745, 3.4686, 0.4553] \\
row 6: [0.0209, 0.2745, 0.342] \\ 
\texttt{------------------------------------------------------------------------------------------------------------------------------------------------------------------------------------------------------------------------------} \\
Final Score: 0.4 $\cdot$ softmax(TF-IDF) + 0.3 $\cdot$ softmax(BM25) + 0.3 $\cdot$ softmax(Dense)
\end{tabular}
\\

\midrule
\textbf{Row Selection} & 
Select Top-40\% → \textbf{Final Selected Row indices: [5.0, 4.0, 1.0]}
\newline
(Selected row final scores: [0.2499, 0.2114, 0.1192]) \\

\midrule
\textbf{Filtered Table (Output)} & 
\begin{tabular}[t]{@{}l@{}}
col: col: \texttt{year | award giving body | category | nominated for | result} \\
row 5: 2012 | 28th pmpc ... for movies | new ... the year | tween ... of 2012 | nominated \\

row 4: 2012 | the young critics circle awards  | best performance | oros | nominated \\

row 1: 2012 | the young critics circle awards  | best performance | oros | nominated \\
\end{tabular}\\

\midrule
\textbf{Real Answer} & \textbf{\texttt{[Refuted]}} \\
\bottomrule
\end{tabular}
\end{table*}


\renewcommand{\arraystretch}{0.95}
\begin{table*}[t]
\centering
\scriptsize
\small
\caption{Example of the ATF filtering process on a AIT-QA instance (index = 77)}
\label{tab:ATF_example of AIT-QA}
\resizebox{\textwidth}{!}{%
\begin{tabular}{p{0.22\linewidth} | p{0.72\linewidth}}
\toprule
\multicolumn{2}{c}{\textbf{Example filtering process of ATF from AIT-QA}} \\
\midrule
\textbf{Question (Input)} & \textit{“Tell me Delta Airline's Total Revenue Per Available Seat Mile in 2016”} \\

\midrule
\textbf{Raw Table (Input)} & 
\begin{tabular}[t]{@{}p{\linewidth}@{}}
col: \texttt{Row Header | Year Ended December 31, - 2017 | Year Ended December 31, - 2016 | Year Ended December 31, - 2015 | Year Ended December 31, - 2014 | Year Ended December 31, - 2013} \\

row 0: Revenue passenger miles | 217,712 | 213,098 | 209,625 | 202,925 | 194,988 \\ 

row 1: Available seat miles | 254,325 | 251,867 | 246,764 | 239,676 | 232,740 \\

...

row 8: Average price per fuel gallon (2) | \$1.68 | \$1.49 | \$1.90 | \$3.47 | \$3.00 \\

row 9: Full-time equivalent employees, end of period | 86,564 | 83,756 | 82,949 | 79,655 | 77,755 \\
\end{tabular} \\

\midrule
\textbf{Predicted Anser Entity} & \textit{number} \\

\midrule
\small
\textbf{Column Description} & 
\begin{tabular}{@{}p{\linewidth}@{}}
\texttt{Row Header}: Categories of operational metrics for Delta Airlines' performance (e.g., Revenue passenger miles, Available seat miles, Passenger mile yield, ...) \\
\texttt{Year Ended December 31, - 2017}: Values for 2017 (e.g., 217{,}712, 254{,}325, 15.99¢, ...) \\
\texttt{Year Ended December 31, - 2016}: Values for 2016 (e.g., 213{,}098, 251{,}867, 15.85¢, ...) \\
\texttt{Year Ended December 31, - 2015}: Values for 2015 (e.g., 209{,}625, 246{,}764, 16.59¢, ...) \\
\texttt{Year Ended December 31, - 2014}: Values for 2014 (e.g., 202{,}925, 239{,}676, 17.22¢, ...) \\
\texttt{Year Ended December 31, - 2013}: Values for 2013 (e.g., 194{,}988, 232{,}740, 16.89¢, ...) \\
\end{tabular} \\

\midrule
\small
\textbf{Column Relevance scores} & 
\begin{tabular}[t]{@{}l@{}}
\textbf{\textit{Score}: [LLM Score, Cosine Score]} \\
Row Header: [0.0, 0.3448] \\ 
Year Ended December 31, - 2017: [0.0, 0.6029] \\ 
Year Ended December 31, - 2016: [1.0, 1.0] \\ 
Year Ended December 31, - 2015: [0.0, 0.0] \\ 
Year Ended December 31, - 2014: [0.0, 0.259] \\ 
Year Ended December 31, - 2013: [0.0, 0.3487] \\ 
\end{tabular}
\\

\midrule
\textbf{Column Clustering} & 
\begin{tabular}[t]{@{}p{\linewidth}@{}}
\textit{Cluster 0}: [\texttt{Year Ended December 31, - 2016}, \texttt{nominated for}] → Selected Cluster → Select ALL\\
\textit{Cluster 1}: [\texttt{Row Header}, \texttt{Year Ended December 31, - 2015}, \texttt{Year Ended December 31, - 2014}, \texttt{Year Ended December 31, - 2013}] → Select \texttt{Year Ended December 31, - 2013}\\
\textit{Cluster 2}: [\texttt{Year Ended December 31, - 2017}] → Select \texttt{Year Ended December 31, - 2017} \\
\end{tabular}
\\

\midrule
\textbf{Essential Columns} & 
Essential Columns: [\texttt{Year Ended December 31, - 2016}] \\

\midrule
\textbf{Column Selection} & 
Essential Columns + Cluster Selected Columns + \texttt{Row Header}\newline
\textbf{Final Selected Columns: [\texttt{'Year Ended December 31, - 2016'}, \texttt{'Year Ended December 31, - 2013'}, \texttt{'Year Ended December 31, - 2017'}, \texttt{Row Header}]} \\

\midrule
\textbf{Flatten Row text} & 
\begin{tabular}[t]{@{}l@{}}
\begin{tabular}[t]{@{}p{\linewidth}@{}}
col: \texttt{Row Header Year Ended December 31, - 2017 Year Ended December 31, - 2016 Year Ended December 31, - 2013} \\

row 0: Revenue passenger miles 217,712 213,098 194,988 \\ 

row 1: Available seat miles 254,325 251,867 232,740 \\

...

row 8: Average price per fuel gallon (2) \$1.68 \$1.49 \$3.00 \\

row 9: Full-time equivalent employees, end of period 86,564 83,756 77,755 \\
\end{tabular} \\ 
\end{tabular}
\\

\midrule
\textbf{Row Hybrid Scores} & 
\begin{tabular}{@{}p{\linewidth}@{}}
\textbf{\textit{Score}: [TF-IDF Score, BM25 Score, Dense Score]} \\
row 0: [0.1078, 0.7664, 0.5872] \\ 
row 1: [0.1432, 0.7396, 0.4875] \\
... \\
row 5: [0.0485, 0.3502, 0.1952] \\
row 6: [0.0, 0.0, 0.2195] \\ 
\texttt{------------------------------------------------------------------------------------------------------------------------------------------------------------------------------------------------------------------------------} \\
Final Score: 0.4 $\cdot$ softmax(TF-IDF) + 0.3 $\cdot$ softmax(BM25) + 0.3 $\cdot$ softmax(Dense)
\end{tabular}
\\

\midrule
\textbf{Row Selection} & 
Select Top-40\% → \textbf{Final Selected Row indices: [4.0, 3.0, 5.0, 0.0]}
\newline
(Selected row final scores: [0.2906, 0.1168, 0.091, 0.0824]) \\

\midrule
\small
\textbf{Filtered Table (Output)} & 
\begin{tabular}[t]{@{}l@{}}
\begin{tabular}[t]{@{}p{\linewidth}@{}}
col: \texttt{Row Header | Year Ended December 31, - 2017 | Year Ended December 31, - 2016 | Year Ended December 31, - 2015 | Year Ended December 31, - 2014 | Year Ended December 31, - 2013} \\

row 0: Revenue passenger miles | 217,712 | 213,098 | 209,625 | 202,925 | 194,988 \\ 

row 1: Available seat miles | 254,325 | 251,867 | 246,764 | 239,676 | 232,740 \\

row 2: Passenger mile yield | 15.99¢ | 15.85¢ | 16.59¢ | 17.22¢ | 16.89¢ \\

row 9: Full-time equivalent ... of period | 86,564 | 83,756 | 82,949 | 79,655 | 77,755 \\
\end{tabular} \\
\end{tabular}\\

\midrule
\textbf{Real Answer} & \textbf{\texttt{['15.74¢']}} \\
\bottomrule
\end{tabular}%
}
\end{table*}

\end{document}